\documentclass{article} 
\usepackage[submission]{SKYLENAGE}

\usepackage{microtype}
\usepackage{hyperref}
\usepackage{url}
\usepackage{booktabs}
\usepackage{graphicx}
\usepackage{subcaption}
\usepackage{caption}
\usepackage{amsmath,amssymb}
\usepackage{xspace}
\usepackage{siunitx}
\usepackage{enumitem}

\IfFileExists{lineno.sty}{%
  \usepackage{lineno} 
  \AtBeginDocument{%
    \ifcolmsubmission
      \linenumbers
    \fi
  }%
}{}

\makeatletter
\@ifpackageloaded{lineno}{\AtBeginDocument{\nolinenumbers}}{}
\makeatother

\definecolor{darkblue}{rgb}{0, 0, 0.5}
\hypersetup{colorlinks=true, citecolor=darkblue, linkcolor=darkblue, urlcolor=darkblue}

\graphicspath{{pdfs/}}
\usepackage[most]{tcolorbox}
\tcbset{colback=white,colframe=black,boxrule=0.6pt,arc=2pt,left=8pt,right=8pt,top=8pt,bottom=8pt}
\newtcolorbox{boxedblock}{}

\newcommand{\ReasonHundred}{\textsc{SKYLENAGE-ReasoningMATH}\xspace}  
\newcommand{\ContestInnovation}{\textsc{SKYLENAGE-MATH}\xspace}        

\title{SKYLENAGE Technical Report: Mathematical Reasoning and Contest-Innovation Benchmarks for Multi-Level Math Evaluation}

\begin{document}

\maketitle

\begin{abstract}

Large language models (LLMs) now perform strongly on many public math suites, yet frontier separation within \emph{mathematics} increasingly suffers from ceiling effects. We present two complementary benchmarks: \textbf{SKYLENAGE\textendash ReasoningMATH}, a 100\textendash item, structure\textendash aware diagnostic set with per\textendash item metadata on length, numeric density, and symbolic complexity; and \textbf{SKYLENAGE\textendash MATH}, a 150\textendash item contest\textendash style suite spanning four stages from high school to doctoral under a seven\textendash subject taxonomy.
We evaluate fifteen contemporary LLM variants under a single setup and analyze subject\,$\times$\,model and grade\,$\times$\,model performance. On the contest suite, the strongest model reaches \textbf{44\%} while the runner-up reaches \textbf{37\%}; accuracy declines from high school to doctoral, and top systems exhibit a doctoral–to–high-school retention near \textbf{79\%}. On the reasoning set, the best model attains \textbf{81\%} overall, and hardest-slice results reveal clear robustness gaps between leaders and the mid-tier.
In summary, we release \ReasonHundred{} and report aggregate results for \ContestInnovation{}; together, \textbf{SKYLENAGE} provides a hard, reasoning\textendash centered and broadly covering math benchmark with calibrated difficulty and rich metadata, serving as a reference benchmark for future evaluations of mathematical reasoning.
\end{abstract}

\section{Introduction}

Large language models (LLMs) are increasingly capable of tackling mathematical problems, a domain that requires not only linguistic fluency but also precise symbolic manipulation and multi-step reasoning. Recent progress has been striking: models can solve grade-school arithmetic word problems with high accuracy and even approach competition-level benchmarks such as the American Invitational Mathematics Examination (AIME)~\cite{codeforcesamerican} and the MATH benchmark (MATH). This trend has elevated mathematics into a central testbed for probing the reasoning abilities of frontier systems.

Yet current evaluations remain limited in several respects. Widely used benchmarks such as Grade School Math 8K (GSM8K)~\cite{cobbe2021training}, MATH~\cite{hendrycks2021measuring}, and AIME slices provide valuable signals but compress heterogeneous capabilities into single aggregate scores. This leads to two key issues. First, {ceiling effects} emerge as strong models saturate existing benchmarks, making it difficult to separate frontier systems. Second, {ability masking} occurs when distinct competencies---for example, resilience at graduate-level problems or strengths in discrete mathematics versus continuous calculus---are hidden behind global averages. Robust evaluation thus requires testbeds that are both {difficult enough} to discriminate at the top end and {structured enough} to reveal fine-grained variation.

Several recent efforts have begun to address these gaps by introducing robustness-oriented datasets (e.g., SVAMP~\cite{patel2021nlp}) or domain-specific stress tests (e.g., Graduate-level Google-Proof Q\&A (GPQA)~\cite{rein2024gpqa} in the sciences). However, in mathematics, there remains a need for benchmarks that simultaneously diagnose structural reasoning ability and capture the breadth of contest-style difficulty spanning multiple academic stages. Such benchmarks should expose subject-level fragmentation, grade-wise resilience, and hardest-slice robustness, all of which are obscured by single-score leaderboards.

\textbf{SKYLENAGE} aims to fill this gap by introducing two complementary benchmarks:
\begin{itemize}[leftmargin=1.2em]
  \item \textbf{\ReasonHundred{}} (100 problems), a reasoning-centric diagnostic set with metadata such as length, numeric density, and symbolic complexity. It emphasizes {structure-first reasoning} over rote computation, enabling analysis of error sensitivity and hardest-quintile retention.
  \item \textbf{\ContestInnovation{}} (150 problems), a contest-style set spanning HS/UG/GR/PhD stages and annotated under a \textbf{seven-subject taxonomy} (Algebra, Calculus, Combinatorics, Geometry, Graph Theory, Number Theory, Probability). Its multi-label design reflects the composite nature of contest problems, and aggregate analyses highlight subject$\times$model and grade$\times$model dynamics.
\end{itemize}

We evaluate 15 contemporary LLM variants under a unified protocol combining chain-of-thought prompting, small-sample self-consistency, standardized answer extraction, and exact-match grading with numeric tolerance. Analyses include subject- and grade-conditioned heatmaps, radar profiles of comparative performance, and structure–performance relationships. Cross-benchmark positioning against public suites such as GSM8K, MATH, AIME, GPQA, and Massive Multitask Language Understanding – Professional (MMLU-Pro) further situates SKYLENAGE in the broader evaluation landscape.

Our study yields several insights: (i) clear tiering among models with stable leader–mid–tail separation; (ii) fragmented leadership across subjects, suggesting opportunities for ensembles; and (iii) steep declines from HS to PhD that sharpen separations at advanced difficulty levels. These findings underscore that single-score leaderboards are insufficient for characterizing mathematical reasoning ability.

\paragraph{Contributions.}
\begin{itemize}[leftmargin=1.2em]
  \item We introduce two complementary math benchmarks—\ReasonHundred{} (structure\textendash aware reasoning diagnostics) and \ContestInnovation{} (contest\textendash style breadth with grade scaling)—that jointly restore headroom and enable fine\textendash grained analysis.
  \item We position \emph{both} tracks as living benchmarks: a frozen static core for comparability and controlled dynamic variants for robustness stress tests in future updates.
  \item We provide comprehensive analyses that uncover subject specialization, grade-band resilience, and robustness to structural difficulty, offering actionable insights for model development and deployment.
\end{itemize}

We publicly release \ReasonHundred{} with metadata and graders, while restricting \ContestInnovation{} to aggregate analyses due to sensitivity. This balanced strategy preserves evaluation value while supporting reproducible research.

\section{Related Work: Mathematical Evaluation of LLMs}

Large language models (LLMs) have rapidly advanced on mathematical problem solving, a capability that stresses both symbolic manipulation and multi-step reasoning. Unlike short-answer NLP tasks, math evaluation demands faithful intermediate reasoning, robust answer extraction, and careful grading protocols. This section reviews core benchmarks, prompting/training methods that drive progress, evaluation methodology, and open challenges relevant to our two benchmarks.

\paragraph{From grade-school arithmetic to Olympiad-level proofs.}
The modern wave of math benchmarks spans from short word problems to competition-style items. \textbf{GSM8K}~\cite{cobbe2021training} established a curated grade-school baseline of high-quality word problems with single-number answers (often paired with free-form rationales in evaluation practice). For broader textual patterns and elementary types, \textbf{AI2 School Math Diverse (ASDiv)}~\cite{miao2021diverse} and the repository-style \textbf{Math Word Problem Solvers (MAWPS)}~\cite{koncel2016mawps} provide diverse or composable math word problems, while \textbf{Algebra Question Answering with Rationales (AQuA-RAT)}~\cite{ling2017program} contributes large-scale algebraic items with natural-language rationales. To probe competition-level reasoning, \textbf{MATH}~\cite{hendrycks2021measuring} offers 12.5K problems across algebra, geometry, number theory, and calculus with step-by-step solutions; community practice further uses small, high-difficulty {AIME/AMC slices} and distilled subsets (e.g., {MATH-500}) as compact stress tests. Robustness-oriented suites such as \textbf{SVAMP}~\cite{patel2021nlp} introduce carefully crafted variants to reduce superficial-cue reliance, while \textbf{MathQA}~\cite{amini2019mathqa} augments AQuA with operation-based program annotations to support interpretable, typed solution programs. Overall, these resources delineate two axes: short word problems that emphasize arithmetic/simple algebra, and contest-style evaluations that stress multi-step symbolic reasoning.

\subsection{Cross-Benchmark Comparisons}
\label{sec:models_refs}

We position our results against widely used public suites to contextualize \textsc{SKYLENAGE-ReasoningMATH} and \textsc{SKYLENAGE-MATH}. We evaluate \textbf{15} contemporary LLM variants spanning proprietary and open-weight families. Many model strings in our harness carry vendor-style build tags (e.g., dates or routing/activation codes such as “A3B’’ or “0709’’). Unless explicitly discussed, we treat these as concrete {variants} within public model {families} (e.g., GPT-5, Gemini~2.5, Qwen3), and we cite the closest official family documentation. We also report scores on {Humanity's Last Exam
(HLE)}, an internal held-out long-form reasoning suite used as a stability anchor across models. Cross-benchmark accuracy is reported for the \textbf{14 models} with public numbers; {GPT-5-Chat-0807} lacks comparable public scores and is omitted from the cross-benchmark table.

\begin{table*}[t]
\centering
\caption{\textbf{Models evaluated and references.} Variants with internal-style tags are mapped to their public family pages for documentation. “Type’’ reflects public claims (dense vs.\ mixture-of-experts, etc.) when available.}
\resizebox{\textwidth}{!}{%
\begin{tabular}{l l l}
\toprule
\textbf{Model string (this paper)} & \textbf{Family / Vendor} & \textbf{Type (public)} \\
\midrule
Kimi\text{-}K2\text{-}Turbo~\cite{team2025kimi} & Kimi K2 / Moonshot AI & MoE (1T total, $\sim$32B active); turbo runtime \\
Gemini2.5\text{-}flash\text{-}0617~\cite{singal2025comparative} & Gemini 2.5 / Google & Proprietary (multimodal) \\
GPT\text{-}5\text{-}Chat\text{-}0807~\cite{hou2025benchmarking} & GPT-5 / OpenAI & Proprietary (routing; chat-tuned) \\
DeepSeek\text{-}V3\text{-}0324~\cite{liu2024deepseek} & DeepSeek-V3 / DeepSeek & Proprietary (reasoning) \\
Llama~4 Maverick~\cite{tang2025efficient} & Llama 4 / Meta & Open-weight family \\
GPT\text{-}5 mini~\cite{hou2025benchmarking} & GPT-5 mini / OpenAI & Proprietary (smaller) \\
GLM\text{-}4.5~\cite{zeng2025glm} & GLM-4.5 / Zhipu AI & Proprietary (multimodal) \\
Gemini2.5\text{-}Pro\text{-}0617~\cite{singal2025comparative} & Gemini 2.5 / Google & Proprietary (Pro tier) \\
GPT\text{-}5\text{-}20250807~\cite{hou2025benchmarking} & GPT-5 / OpenAI & Proprietary (flagship) \\
GPT\text{-}oss\text{-}120b~\cite{hou2025benchmarking} & GPT-OSS / OpenAI & Open-weight (reasoning) \\
Ernie\text{-}4.5\text{-}424B\text{-}A47B~\cite{sun2021ernie} & ERNIE 4.5 / Baidu & Proprietary (multimodal) \\
DeepSeek\text{-}V3.1~\cite{guo2025deepseek} & DeepSeek-V3.1 / DeepSeek & Proprietary (reasoning) \\
DeepSeek\text{-}R1\text{-}0528~\cite{guo2025deepseek} & DeepSeek-R1 / DeepSeek & Distilled reasoning \\
Grok\text{-}4\text{-}0709~\cite{bussaja2025analyzing} & Grok 4 / xAI & Proprietary (frontier) \\
Qwen3\text{-}235B\text{-}A22B\text{-}2507~\cite{yang2025qwen3} & Qwen3 Family / Qwen Team & Open-weight family (235B variants) \\
\bottomrule
\end{tabular}%
}
\label{tab:models-and-refs}
\end{table*}

\begin{table*}[t]
\centering
\caption{\textbf{Cross-benchmark accuracy results (sorted by macro mean).} Rows are models and columns are benchmarks. “Mean’’ is the macro average over available benchmarks for that model (missing entries are ignored). A dash (\texttt{--}) denotes a missing evaluation; per-column best is bolded.}
\resizebox{\textwidth}{!}{%
\begin{tabular}{lccccccc}
\toprule
\textbf{Model} & AIME25 & AIME24 & MATH-500 & HLE & GPQA & MMLU-Pro & \textbf{Mean} \\
\midrule
GPT-5-20250807       & \textbf{99.6} & \textbf{94.2} & \textbf{99.4} & \textbf{26.5} & 85.4 & \textbf{87.1} & \textbf{82.0} \\
Grok-4-0709          & 92.7 & 90.6 & 96.2 & 23.9 & \textbf{87.7} & 86.6 & 79.6 \\
Gemini2.5-Pro-0617   & 88.0 & 87.5 & 96.7 & 21.1 & 84.4 & 86.2 & 77.3 \\
GPT-5 mini           & 90.7 & 90.8 & 94.8 & 19.7 & 82.8 & 83.7 & 77.1 \\
DeepSeek-R1-0528     & 87.5 & 91.4 & 98.3 & 14.9 & 81.3 & 84.9 & 76.4 \\
GPT-oss-120b         & 97.9 & 93.1 & --   & 18.5 & 78.2 & 80.8 & 73.7 \\
Qwen3-235B-A22B-2507 & 81.5 & 85.7 & 90.2 & 15.0 & 79.0 & 84.3 & 72.6 \\
DeepSeek-V3.1        & 88.4 & 93.1 & --   & 13.0 & 77.9 & 85.1 & 71.5 \\
GLM-4.5              & 73.7 & 68.8 & 98.2 & 12.2 & 78.2 & 83.5 & 69.1 \\
Kimi-K2-Turbo        & 57.0 & 70.0 & 97.1 &  7.0 & 76.6 & 82.4 & 65.0 \\
Gemini2.5-flash-0617 & 72.0 & --   & 93.2 &  5.1 & 68.3 & 80.9 & 63.9 \\
DeepSeek-V3-0324     & 51.3 & 59.4 & 94.2 &  5.2 & 65.5 & 81.9 & 59.6 \\
Ernie-4.5-424B-A47B  & 35.1 & 54.8 & 96.4 & --   & --   & 58.8 & 61.3 \\
Llama 4 Maverick     & 19.3 & 25.2 & 85.2 &  4.8 & 67.1 & 80.9 & 47.1 \\
\bottomrule
\end{tabular}%
}
\label{tab:cross-benchmark}
\end{table*}

\vspace{0.4em}
\begin{figure}[t]
  \centering
  \includegraphics[width=\linewidth]{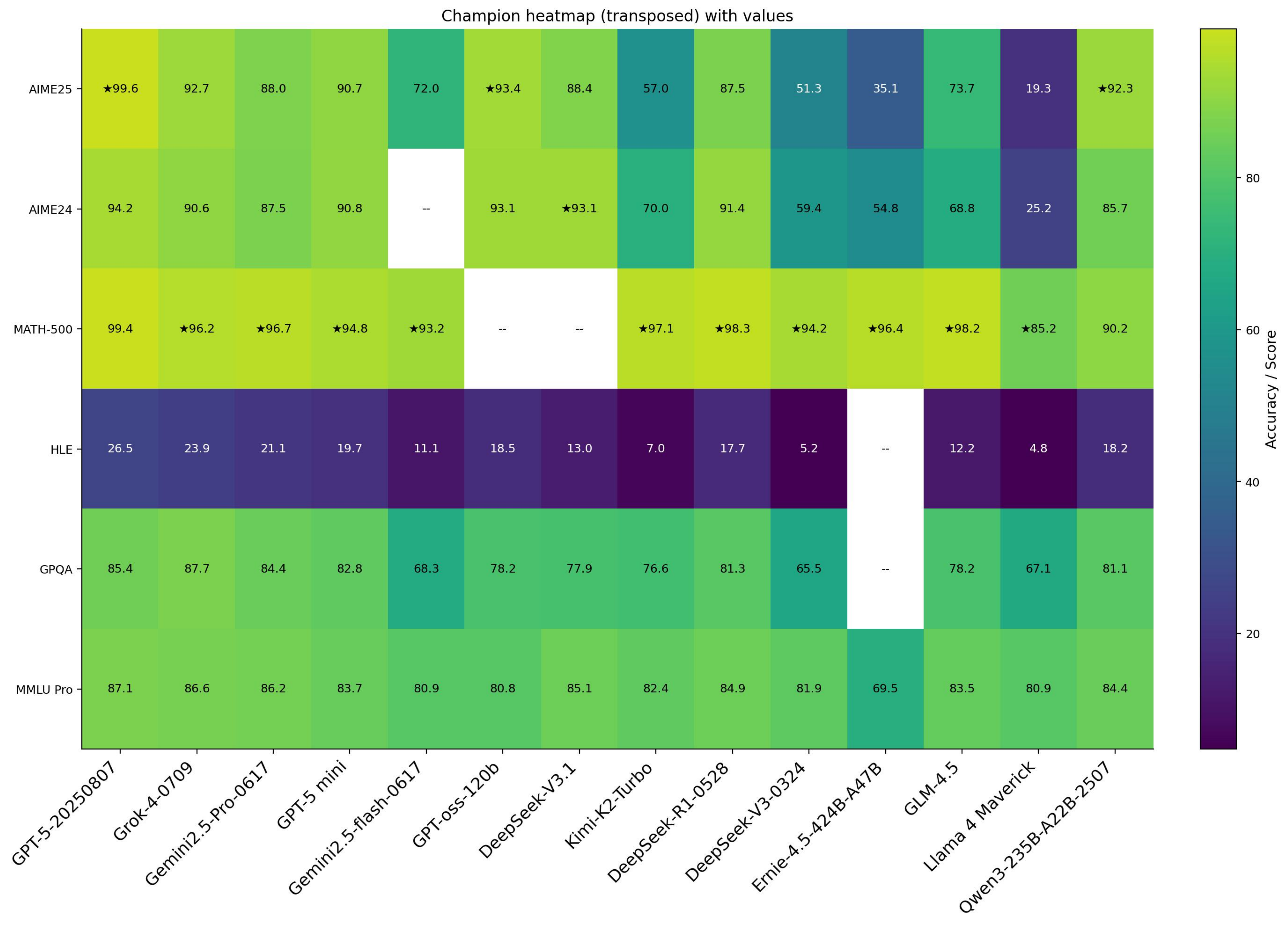}
  \caption{\textbf{Champion heatmap across benchmarks (transposed).} Rows are benchmarks and columns are models. Each cell shows the accuracy; stars mark the per-benchmark champion (ties allowed).}
  \label{fig:champion-heatmap}
\end{figure}

\paragraph{Macro structure and separation.}
The macro mean places {GPT-5-20250807} first at \textbf{82.0}, with a \textbf{+2.4}-point advantage over {Grok-4-0709} (\textbf{79.6}) and a \textbf{+5.6}-point edge over {Gemini2.5-Pro-0617} (\textbf{77.3}); relative to the \#5 model ({DeepSeek-R1-0528}, \textbf{76.4}), the margin is \textbf{+5.6} points (\(\sim\)\textbf{+7.3\%} relative). These gaps persist despite saturation on some columns, indicating a stable top tier rather than a single outlier model.

\paragraph{Discriminative power and ceiling effects.}
Benchmarks differ notably in spread. On \textbf{AIME25}, the cohort spans \textbf{80.3} points (from \textbf{99.6} to \textbf{19.3}); on \textbf{AIME24}, the range is \textbf{69.0} (\textbf{94.2} to \textbf{25.2}). In contrast, \textbf{MATH-500} compresses to a \textbf{14.2}-point band (\textbf{99.4} to \textbf{85.2}), with a top–runner gap of only \textbf{2.7} (99.4 vs.\ 96.7). Knowledge-heavy suites show intermediate dispersion: \textbf{GPQA} ranges \textbf{22.2} (87.7–65.5) and \textbf{MMLU-Pro} \textbf{28.3} (87.1–58.8). The long-form anchor \textbf{HLE} remains intentionally hard (range \textbf{21.7}, 26.5–4.8). Together, these ranges quantify where frontier systems still separate: AIME-style contests remain sensitive, MATH-500 strongly saturates, and GPQA/MMLU-Pro capture breadth beyond math.


\paragraph{Agreement with long-form reasoning and rotation of champions.}
\textbf{HLE} emphasizes sustained multi-step derivations; the champion heatmap in Fig.~\ref{fig:champion-heatmap} shows rotation of leaders across suites (e.g., {GPT-5-20250807} leads \textbf{4/6} columns, {Grok-4-0709} leads \textbf{GPQA}). Quantitatively, Supplementary~\ref{app:hle} reports strong alignment between HLE and \textbf{AIME25}/\textbf{GPQA} (highest Pearson \(r\)), moderate alignment for \textbf{AIME24}/\textbf{MMLU-Pro}, and weaker alignment for \textbf{MATH-500}, consistent with the latter’s ceiling compression. The combination of high AIME sensitivity and GPQA knowledge grounding explains why top–runner margins are relatively larger on HLE and AIME than on MATH-500.

\section{Dataset Construction}

\subsection{\ReasonHundred: Design Goals, Sources, and Anti-Contamination}
\label{sec:reason100_construction}

\begin{figure}[htbp]
  \centering
  \includegraphics[width=\linewidth]{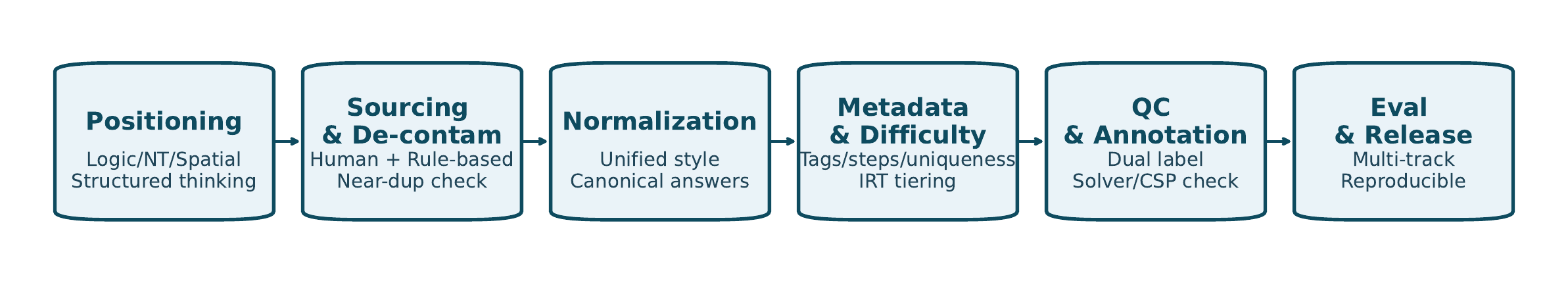}
    \caption{\textbf{\ReasonHundred{} construction pipeline.} Our construction pipeline begins with a three-source intake—human authoring, rule-based generation, and structure-preserving rewrites—followed by multi-pass anti-contamination checks at the string, semantic, and template levels. We then perform style and format normalization, carry out bilingualization to ensure parity across languages, and add minimal process-hook annotations to enable step checks. Quality control is conducted with solver and simulator validation, after which we run a small pilot for difficulty calibration. Finally, we freeze the set for release.}
  \label{fig:reasoning_flow}
\end{figure}
Guided by the above design principles—and to ensure structure-first reasoning, contamination control, and reproducibility—we adopt a staged construction workflow (see Fig.~\ref{fig:reasoning_flow}). Specifically, the pipeline proceeds as follows:

\paragraph{Design goals}
\ReasonHundred{} targets {structure-first reasoning} instead of heavy computation:
(1). Emphasize \textbf{logic/constraint puzzles}, \textbf{number-theoretic/combinatorial constructions}, and \textbf{spatial/geometry intuition} over rote algebra;
(2). Require {decomposable} reasoning with {verifiable intermediate assertions};
(3). Reduce exposure bias by prioritizing {low-frequency} patterns in common pretraining corpora (less templateable, less likely memorized). This design addresses known weaknesses of prior math sets (data leakage, answer-only scoring, format fragility) and shifts the focus from “getting the final number’’ to “reasoning correctly.’’ \textit{(details aligned with our internal design notes).} 

\paragraph{Sourcing and normalization}
We combine {human-authored}, {rule-generated}, and {structure-preserving rewrites}: (1). Authors seed puzzle skeletons and spatial scenarios; (2). Rule-based generators instantiate constraints (entity names, graph sizes) to diversify surface forms without changing solution structure; (3). Bilingual normalization (CN$\leftrightarrow$EN) ensures term consistency and difficulty parity. All items are rewritten for a uniform style (concise statements, explicit constraints) and answerability without diagrams.

\paragraph{Anti-contamination (multi-pass)}
To mitigate train–test leakage, every candidate passes (1). \textbf{string-level} n-gram fingerprinting, (2). \textbf{semantic-level} embedding nearest-neighbor search, and (3). \textbf{template-level} paraphrase detection. High-similarity candidates are {rewritten} or {removed}. We maintain per-item \texttt{hash\_id} and release an aggregate “suspected overlap’’ statistic with our public split. Items are designed to avoid high-frequency classroom templates and to diminish “prompt overfitting’’ to a few-shot layout.  This aligns with our positioning of focusing on logic/space reasoning beyond standard exam styles. 

\paragraph{Metadata and controllable difficulty.}
Each item is tagged with: (1). {subjects} (7-way forced taxonomy: Algebra, Calculus, Combinatorics, Geometry, Graph Theory, Number Theory, Probability; multi-label permitted); (2). {structural features} (length, numeric-token density, symbolic-token count, constraint count, branching factor); (3). {process hooks} (required intermediate assertions, e.g., adjacency tables for logic puzzles; cut/merge invariants for spatial items). We calibrate difficulty with a small pilot-of-models and human raters; a composite difficulty score is derived via rank aggregation over success rates and estimated step depth.

\paragraph{Step-checkable annotations.}
Beyond final answers, we store minimal {checkable} invariants: e.g., for constraint puzzles, a canonical assignment table; for spatial tasks, a vertex/edge transform log; for number theory, key lemmas (parity/modulo) to verify consistency of the chain. These support {process consistency} checks alongside exact-match grading.

\paragraph{Quality control (QC)}

We apply double annotation with arbitration; logic items are validated by a constraint satisfaction problem (CSP) / satisfiability modulo theories (SMT) solver for {uniqueness} and {consistency}; spatial items are replayed with a simple simulator to confirm the stated invariant; bilingual parity is verified by back-translation and spot-checking of {model agreement} (prediction consistency across CN/EN).

\subsection{\ContestInnovation: Curation Protocol and Dataset Characteristics (150)}
\label{sec:ci150_construction}

\paragraph{Curation (expert-driven)}
\ContestInnovation{} comprises 150 {contest-style} problems authored/selected by subject experts, stratified by four stages (HS/UG/GR/PhD). To protect sensitive/licensed content and maintain future evaluation value, raw items are not released; we instead publish aggregate analyses and artifacts that reveal {distributions} without reconstructing items.

\paragraph{Coverage and stratification.}
Each item carries one or more of the seven subjects with a {forced} taxonomy (same as \ReasonHundred{}), and belongs to one of four difficulty stages (HS/UG/GR/PhD). The set intentionally {mixes} single-skill questions and cross-topic composites (e.g., Algebra+Geometry) to reflect contest reality.

\paragraph{Answer types and grading policy.}
Items are auto-graded to a canonical final form (integer, fraction, set, or symbolic expression with normalization). Multi-label analysis uses {full-credit} per tagged subject to avoid fractional bookkeeping. Numeric tolerance ($10^{-6}$) is applied when a problem explicitly accepts floating outputs; otherwise, exact-form matching is enforced with a normalizer (common radicals/fractions/ordering).

\paragraph{Dataset-facing highlights}
We present:
(1). \textbf{stage distributions} (HS/UG/GR/PhD) and accuracy gradients;
(2). \textbf{subject coverage} heatmaps and {per-subject champions};
(3). \textbf{cross-subject composites} (share of multi-label items) to expose structural coupling;
(4). \textbf{answer-type mix} (numeric vs.\ symbolic) and its impact on accuracy;
(5). \textbf{subject$\times$stage} performance surfaces to diagnose where gaps widen (e.g., discrete domains at GR/PhD).
These views emphasize {dataset characteristics} without revealing items. 

\paragraph{Rationale.}
The curation targets “contest innovation’’: multi-lemma reasoning, diagram-free geometry, constructive number theory, and discrete structures. The four-stage stratification ensures that top-tier separations appear where prior public sets saturate, while the subject taxonomy enables actionable routing/ensembles in downstream analysis.

\section{Evaluation Protocol}
\begin{itemize}
    \item Prompting and decoding: We use Chain-of-Thought prompting to elicit stepwise reasoning. These practices are widely reported to yield large and consistent gains on arithmetic and symbolic reasoning benchmarks such as GSM8K and related suites~\cite{wei2022chain,wang2022self}. Decoding hyperparameters (temperature, top-p) are kept identical within cost tiers.
    \item Answer extraction and normalization: We standardize final answers via regex templates for integers, fractions, sets, and symbolic forms. For floating-point results, a numeric tolerance of $10^{-6}$ is applied unless the item specifies an exact form. Units and common equivalent formats are normalized.
    \item Grading: Binary exact match (1/0) is applied after normalization. For \ContestInnovation{}, items may carry multiple subject tags; analysis uses a {multi-label, full-credit} convention (an item contributes fully to every tagged subject) to avoid splitting credit.
    \item Harness and fairness: All models are queried by the same harness with identical prompts and extraction rules. Rate limits and batching are controlled to reduce variance. Seeds and decode parameters are recorded for reproducibility.
    \item Process-aware extensions: Future releases will pair exact-match grading with process-aware checks (step validity, constraint fidelity, verifier agreement), computed from minimal per-item hooks, to yield complementary CoT-based scores.
\end{itemize}

\section{Dataset Analysis}
\subsection{Dataset I: \ReasonHundred (100 reasoning problems)}\label{sec:datasetA}

\subsubsection{Overall results and hardest-slice accuracy}

\begin{figure}[t]
  \centering
  \includegraphics[width=\linewidth]{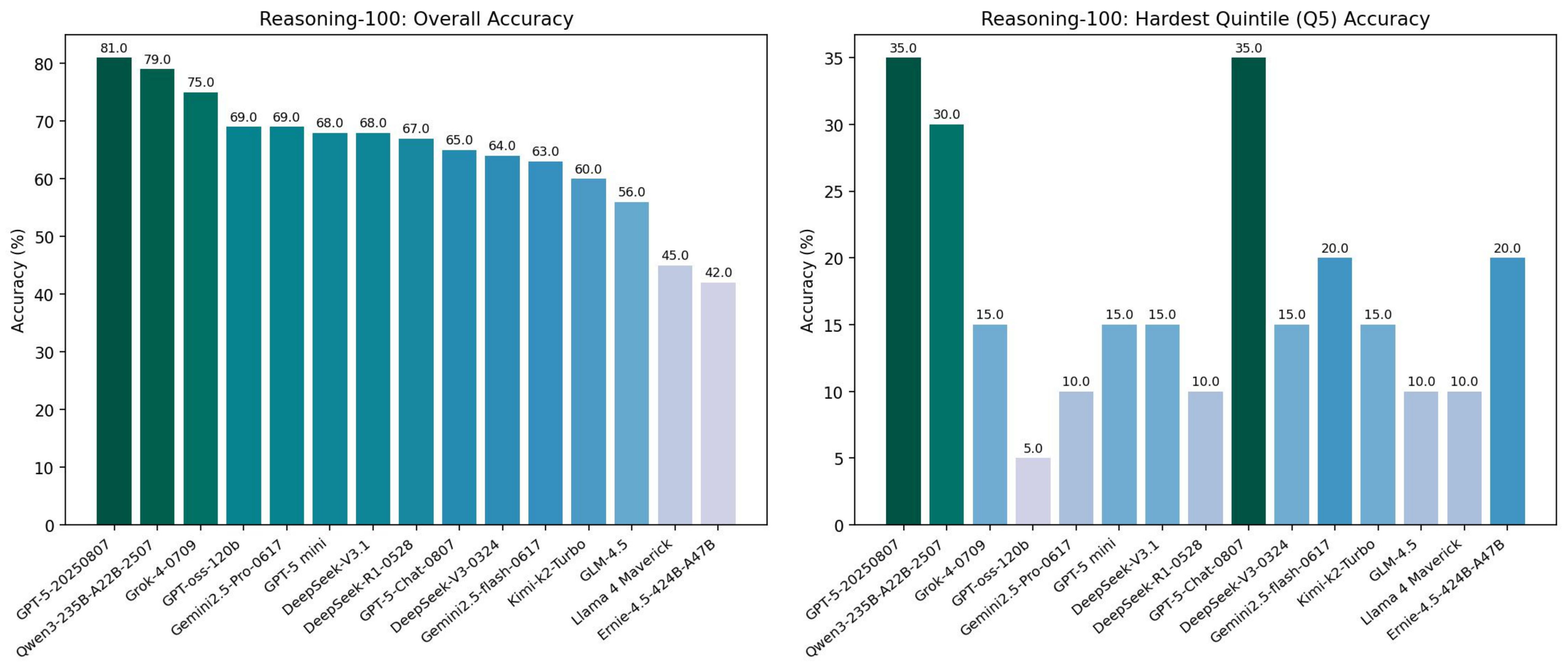}
    \caption{\textbf{Reasoning-100 overview.} {Left}: overall accuracy (sorted, \%). {Right}: accuracy on the {hardest quintile} (Q5). \textit{GPT-5-20250807} reaches \textbf{81\%}, \textit{Qwen3-235B-A22B-2507} follows closely at \textbf{79\%}, and \textit{Grok-4-0709} at \textbf{75\%}. Against the tail, the margin is \textbf{+44.6\%} vs.\ \textit{GLM-4.5} (56\%), \textbf{+80.0\%} vs.\ \textit{Llama 4 Maverick} (45\%), and \textbf{+92.9\%} vs.\ \textit{Ernie-4.5-424B-A47B} (42\%). Top-5 overall (descending): \textit{GPT-5-20250807} (81), \textit{Qwen3-235B-A22B-2507} (79), \textit{Grok-4-0709} (75), \textit{GPT-oss-120b} (69), \textit{Gemini2.5-Pro-0617} (69). On the hardest quintile, \textit{GPT-5-Chat-0807} leads at \textbf{35\%}; \textit{GPT-5-20250807} and \textit{Qwen3-235B-A22B-2507} follow at \textbf{30\%}.}
  \label{fig:reason100-bars}
\end{figure}

\noindent\textit{Analysis.}
We interpret accuracy not as an endpoint but as evidence of process stability under increasing structural load(see Fig.~\ref{fig:reason100-bars}). \ReasonHundred{} separates systems not only by overall accuracy but by {stability under difficulty}. While the flagship’s 81\% exceeds Qwen’s 79\% by +2.5\% (relative to 79) and Grok’s 75\% by +8.0\% (relative to 75), the high-difficulty slice magnifies gaps: the flagship’s Q5 retention is \(\approx\)0.37 and Qwen’s is \(\approx\)0.38, which are \textbf{+38.6\%} and \textbf{+42.5\%} higher than Grok’s \(\approx\)0.27 (computed relative to 0.27). Versus a 69\% mid-tier with \(\le\)10\% on Q5 (retention \(\le\)0.145), the flagship’s 0.37 retention represents a \textbf{+155\%} improvement (relative to 0.145). Hence, among models with similar top-line scores, Q5 retention provides a sharper discriminator of {plan integrity} under branching constraints. See Supplementary~\ref{app:reason100-case-cos-sum} for a structure-first trigonometry case that illustrates how a short nonnegativity bound plus a constructive witness outperforms long formula-chains and exposes typical failure modes (bound--attainment confusion, constraint drops, and identity drift).

\subsubsection{Subject- and difficulty-wise profiling}

\begin{figure}[t]
  \centering
  \includegraphics[width=\linewidth]{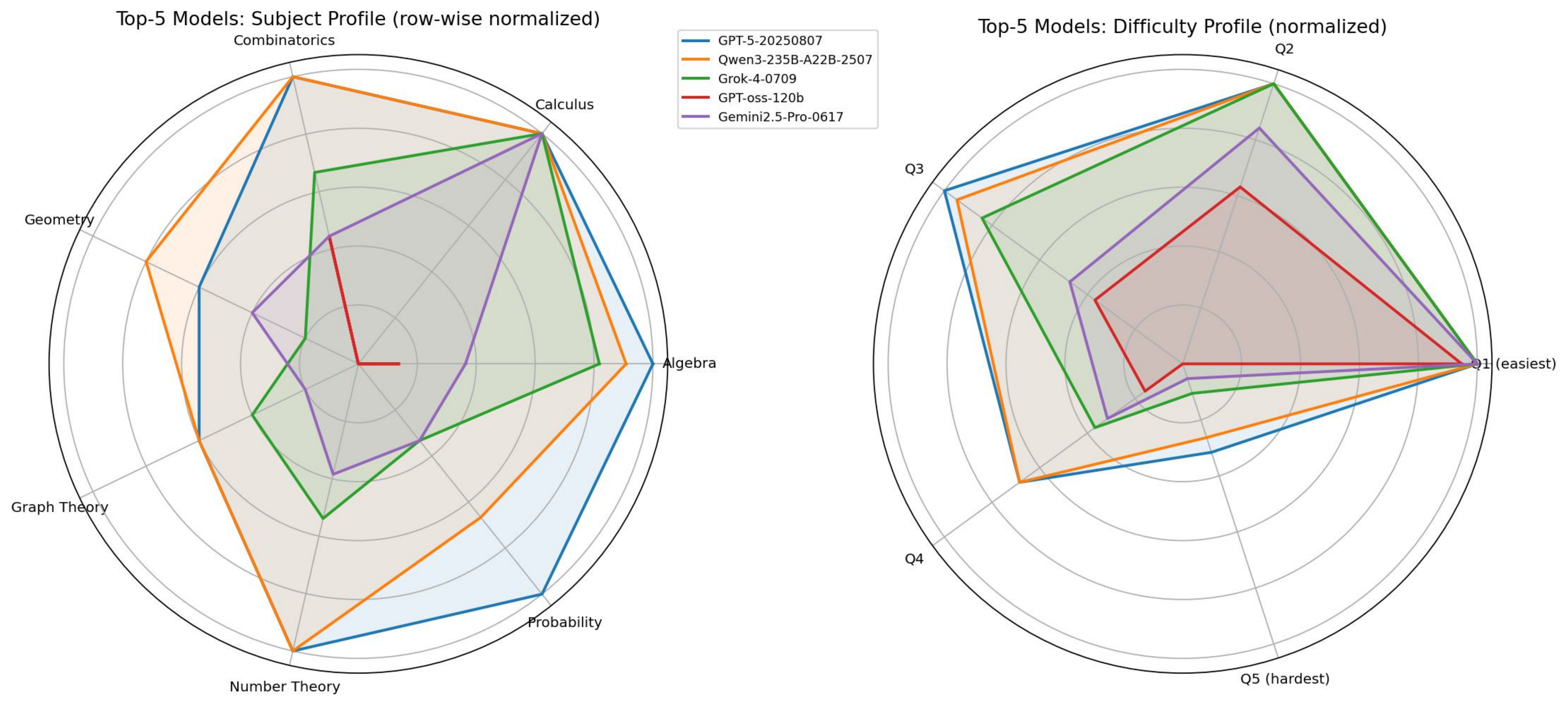}
    \caption{\textbf{Top-5 profiles.} {Left}: subject radar under the \textbf{seven} categories. {Right}: difficulty radar by quintiles Q1--Q5. The flagship dominates discrete-heavy categories: {Combinatorics} \textbf{92.9\%} vs.\ Grok \textbf{71.4\%}, {Probability} \textbf{83.3\%} vs.\ \textbf{50.0\%}, and {Number Theory} \textbf{81.0\%} vs.\ \textbf{52.4\%}. Qwen nearly matches the flagship in most subjects and even surpasses it in {Geometry} (\textbf{75.0\%} vs.\ \textbf{68.8\%}).  In {Calculus}, leaders cluster near \textbf{77.8\%}; Graph Theory shows a notable outlier at \textbf{100\%} (Llama 4 Maverick, likely small-$n$). All models degrade from Q1$\rightarrow$Q5. The flagship and Qwen retain \textbf{37–38\%} of their baseline, vs.\ Grok’s \textbf{20\%} and GPT-oss-120b’s \(\le\)\textbf{15\%}. }
  \label{fig:reason100-radars}
\end{figure}

\noindent\textit{Analysis.}
Subject profiles reveal {rotating leadership} with large \% gaps in discrete domains (see Fig.~\ref{fig:reason100-radars}): the flagship’s Combinatorics 92.9\% exceeds Grok’s 71.4\% by \textbf{+30.1\%} (21.5/71.4), Probability 83.3\% exceeds 50.0\% by \textbf{+66.6\%}, and Number Theory 81.0\% exceeds 52.4\% by \textbf{+54.6\%}. Conversely, Qwen’s Geometry 75.0\% exceeds the flagship’s 68.8\% by \textbf{+9.0\%}. On difficulty, the Top-2’s Q5 retention (37–38\%) is \textbf{+85–90\%} higher than Grok’s 20\% and \textbf{+146–153\%} higher than a \(\le\)15\% mid-tier (all relative to the comparator), indicating that discrete strengths translate into measurably slower degradation as problems become more compositional.

\subsubsection{Subject \texorpdfstring{$\times$}{x} model heatmap}

\begin{figure}[t]
  \centering
  \includegraphics[width=1\linewidth]{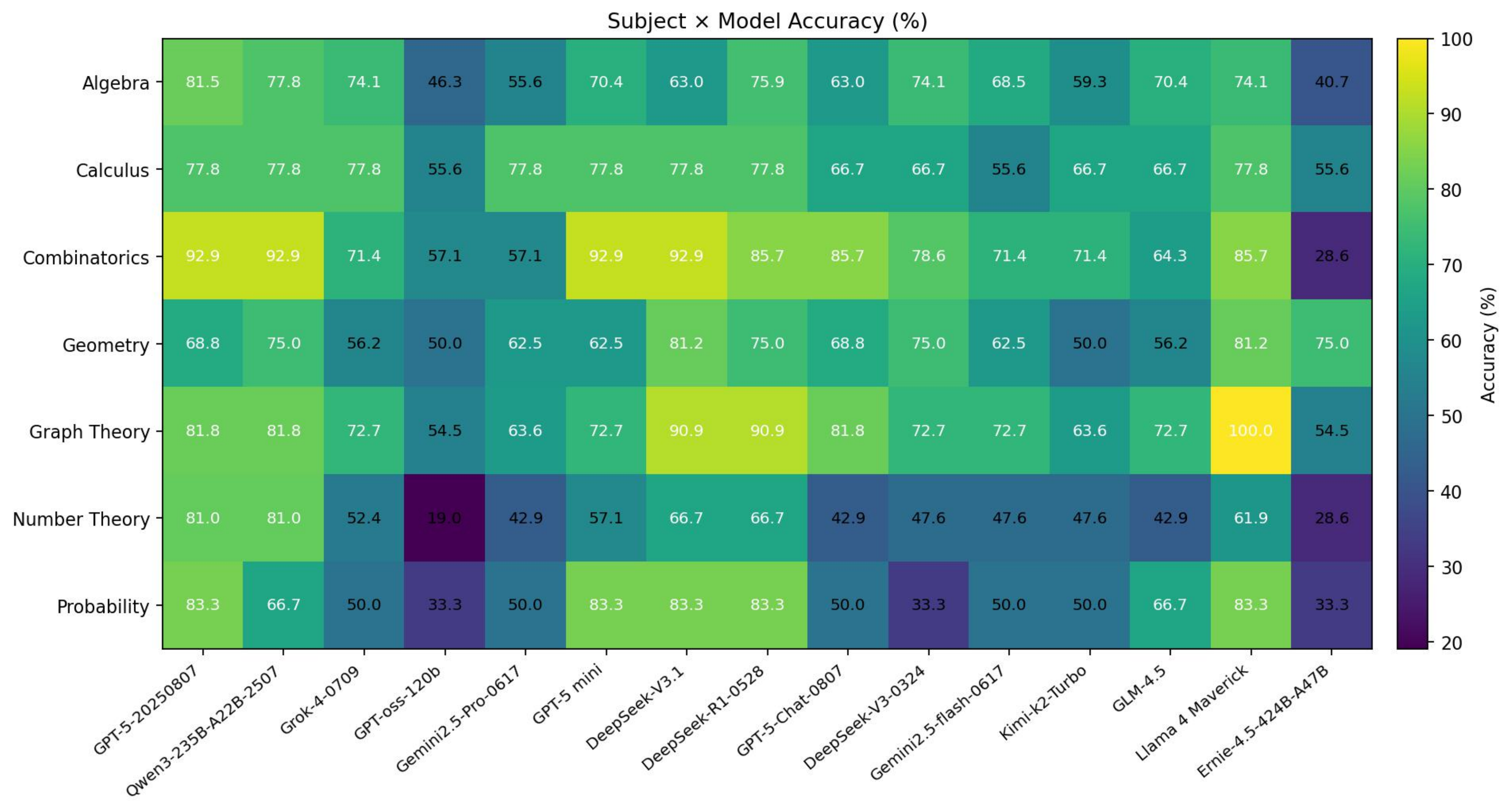}
    \caption{\textbf{Subject $\times$ model accuracy heatmap} (\%). Seven-subject taxonomy. Darker = higher accuracy. Qwen3-235B-A22B-2507 nearly matches the flagship in most subjects and even surpasses it in {Geometry}.}
  \label{fig:reason100-heat}
\end{figure}

\noindent\textit{Analysis.}
Complementarity is quantifiable: Qwen’s Geometry lead of +6.2 points equates to \textbf{+9.0\%} relative to the flagship’s 68.8, while the flagship’s Probability edge (83.3 vs.\ 66.7) is \textbf{+24.9\%} relative to 66.7 and its Number Theory edge (81.0 vs.\ 52.4) is \textbf{+54.6\%}(see Fig.~\ref{fig:reason100-heat}). A per-subject oracle that selects the best family per cell would thus harvest multiple \% gains over any single model; gains are largest where the leading margin exceeds 20\% relative (discrete-heavy cells) and smallest where leaders cluster (e.g., Calculus).


\subsubsection{Structure vs.\ performance}

\begin{figure}[htbp]
  \centering
  \makebox[\linewidth][c]{%
    \includegraphics[width=1.1\linewidth]{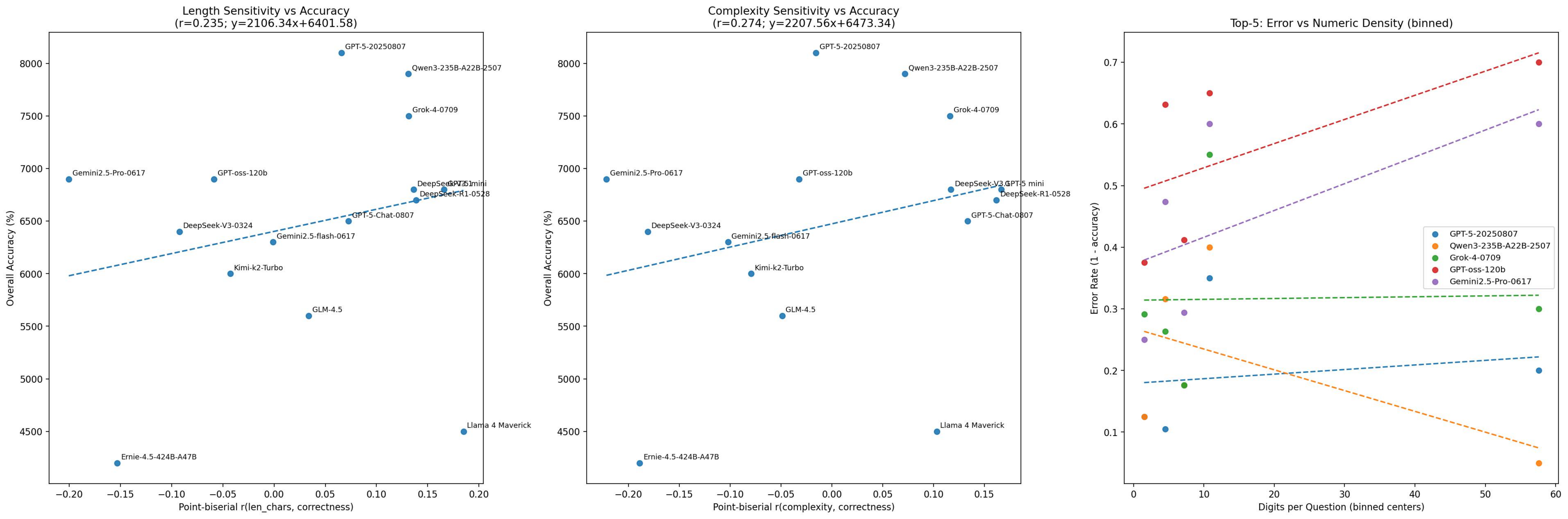}%
  }
\caption{\textbf{Structure--performance relationships.} {Left}: sensitivity to length vs.\ accuracy. {Middle}: complexity sensitivity vs.\ accuracy. {Right}: error vs.\ numeric density (top-5 models). Length and complexity sensitivities show weak positive correlations (\(r\approx0.2\)). Numeric density is sharper:  \textit{GPT-oss-120b} errors surge (\(+92\%\)), \textit{Gemini2.5-Pro-0617} rises \(\sim 30\%\), flagship \textit{GPT-5-20250807} only \(\sim 18\%\), Grok nearly flat, and \textit{Qwen} trends negative (errors decline as digits grow).}
  \label{fig:reason100-scatters}
\end{figure}


\noindent\textit{Analysis.}
Length and symbolic complexity show only weak positive association with errors ($r{\approx}0.2$), suggesting that mere sequence size or token variety is not the principal driver of failure. By contrast, \emph{numeric density} (share of digits in the prompt) consistently separates families: an open-weight 120B variant exhibits steep error inflation as digits increase (on the order of ${\sim}{+}90\%$ across density bins), a strong proprietary baseline inflates more modestly (${\sim}{+}30\%$), while the flagship shows only a mild rise (${\sim}{+}18\%$)(see Fig.~\ref{fig:reason100-scatters}). Qwen trends close to flat or slightly negative. Combined with Q5 retention, this indicates that arithmetic normalization and digit handling—rather than sheer length—drive the largest relative differences at the frontier and should be a priority for targeted finetuning and decoding policies.

\subsubsection{Hardest items (diagnostics)}

\begin{table}[t]
  \centering
  \caption{\textbf{Top-10 hardest items} (lowest mean accuracy). ``Len'': characters; ``Digits'': number tokens; ``Symbols'': math tokens. The hardest 10 items are dominated by \textbf{Algebra} (6/10) and \textbf{Number Theory} (3/10). Mean accuracies (\(\le 11.8\%\)) are $\sim$83\% below mid-cluster and $\sim$85\% below the flagship. }
  \resizebox{\linewidth}{!}{%
  \begin{tabular}{l l r r r r r}
    \toprule
    Item & Subject(s) & Len & Digits & Symbols & Complexity $z$ & Mean Acc (\%) \\
    \midrule
    Q056 & Number Theory                      & 20  & 11 & 0  & $-0.48$ & \textbf{0.0} \\
    Q084 & Algebra                            & 20  & 11 & 0  & $-0.48$ & \textbf{5.9} \\
    Q092 & Number Theory                      & 23  & 4  & 0  & $-0.71$ & \textbf{5.9} \\
    Q006 & Algebra                            & 143 & 15 & 1  & $+0.27$ & \textbf{5.9} \\
    Q049 & Calc./NumTh./Prob.                 & 91  & 9  & 3  & $-0.17$ & \textbf{5.9} \\
    Q009 & Algebra                            & 101 & 14 & 1  & $+0.03$ & \textbf{5.9} \\
    Q040 & Algebra                            & 133 & 8  & 8  & $+0.06$ & \textbf{5.9} \\
    Q077 & Geometry/Graph Theory              & 244 & 9  & 10 & $+0.66$ & \textbf{11.8} \\
    Q058 & Algebra                            & 49  & 5  & 6  & $-0.48$ & \textbf{11.8} \\
    Q033 & Algebra/Geometry                   & 297 & 8  & 13 & $+0.92$ & \textbf{11.8} \\
    \bottomrule
  \end{tabular}}
  \label{tab:reason100-hardest}
\end{table}


\noindent\textit{Analysis.}
The hardest slice concentrates in Algebra and Number Theory with compact prompts but high digit share, plus a few long, symbol-rich composites (Table~\ref{tab:reason100-hardest}). These two morphologies align with the dominant failure modes: (i) arithmetic/normalization slips on digit-dense short items and (ii) step drift on long multi-label composites. Mean accuracies $\le$11.8\% are $\sim$85\% below the flagship’s overall level, pinpointing where process checks (algebraic normalizers, simple verifiers) could yield the largest fractional gains.

\subsection{Dataset II: \ContestInnovation{} (150 contest-style problems across HS\texorpdfstring{$\rightarrow$}{→}PhD)}\label{sec:datasetB}

\subsubsection{Composition and Meta-Overview}

\begin{figure}[htbp]
  \centering
  \makebox[\linewidth][c]{%
    \includegraphics[width=1.1\linewidth]{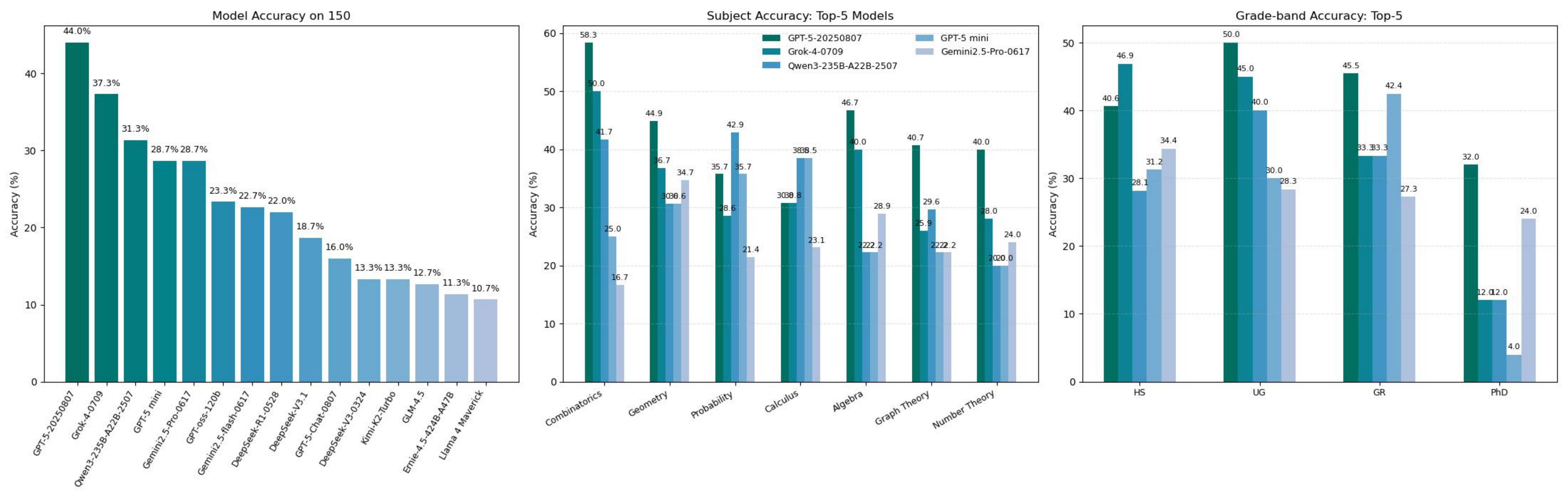}%
  }
    \caption{\textbf{Meta overview.}
  {Left}: overall accuracy for 14 models (\%).
  {Middle}: subject-wise accuracy (top-5 models).
  {Right}: grade-band accuracy (High School (HS) / Undergraduate (UG) / Graduate (GR) / Doctoral (PhD)). The top performer (\textit{GPT-5-20250807}) achieves \textbf{44.0\%}, leading the runner-up (\textit{Grok-4-0709}, 37.3\%). Qwen3-235B-A22B-2507 follows at 31.3\%, close to the second tier (\textit{GPT-5 mini}, 28.7\%; \textit{Gemini2.5-Pro-0617}, 28.7\%). This establishes a three-tier separation: (i) leaders above 35\%, (ii) mid-cluster around 22–31\%, and (iii) tail under 20\%. The gap between the leader and the weakest model (\textit{Llama 4 Maverick}, 10.7\%) is \textbf{+310\%} relative.}
  \label{fig:bars150}
\end{figure}

\noindent\textit{Analysis.}
Contest-style performance emphasizes multi-lemma planning and symbolic canonicalization, making relative gaps more diagnostic than absolute scores. Relative gaps widen in contest settings. The flagship’s 44.0\% exceeds Grok’s 37.3\% by \textbf{+17.9\%} and Qwen’s 31.3\% by \textbf{+40.6\%}. The leader–tail spread (44.0 vs.\ 10.7) is \textbf{+311\%} relative to 10.7. Across grades, PhD’s 14.1\% trails HS’s 26.3\% by \textbf{\( -46.4\%\)}(see Fig.~\ref{fig:bars150}). The three-tier landscape persists across resamplings and decoding settings, indicating that multi-lemma planning and canonicalization pressures amplify separations that saturate on public sets. Grade scaling steepens these gaps further (see Fig.~\ref{fig:heat150}).

\subsubsection{Subject accuracy and champions}

\begin{figure}[t]
  \centering
  \includegraphics[width=\linewidth]{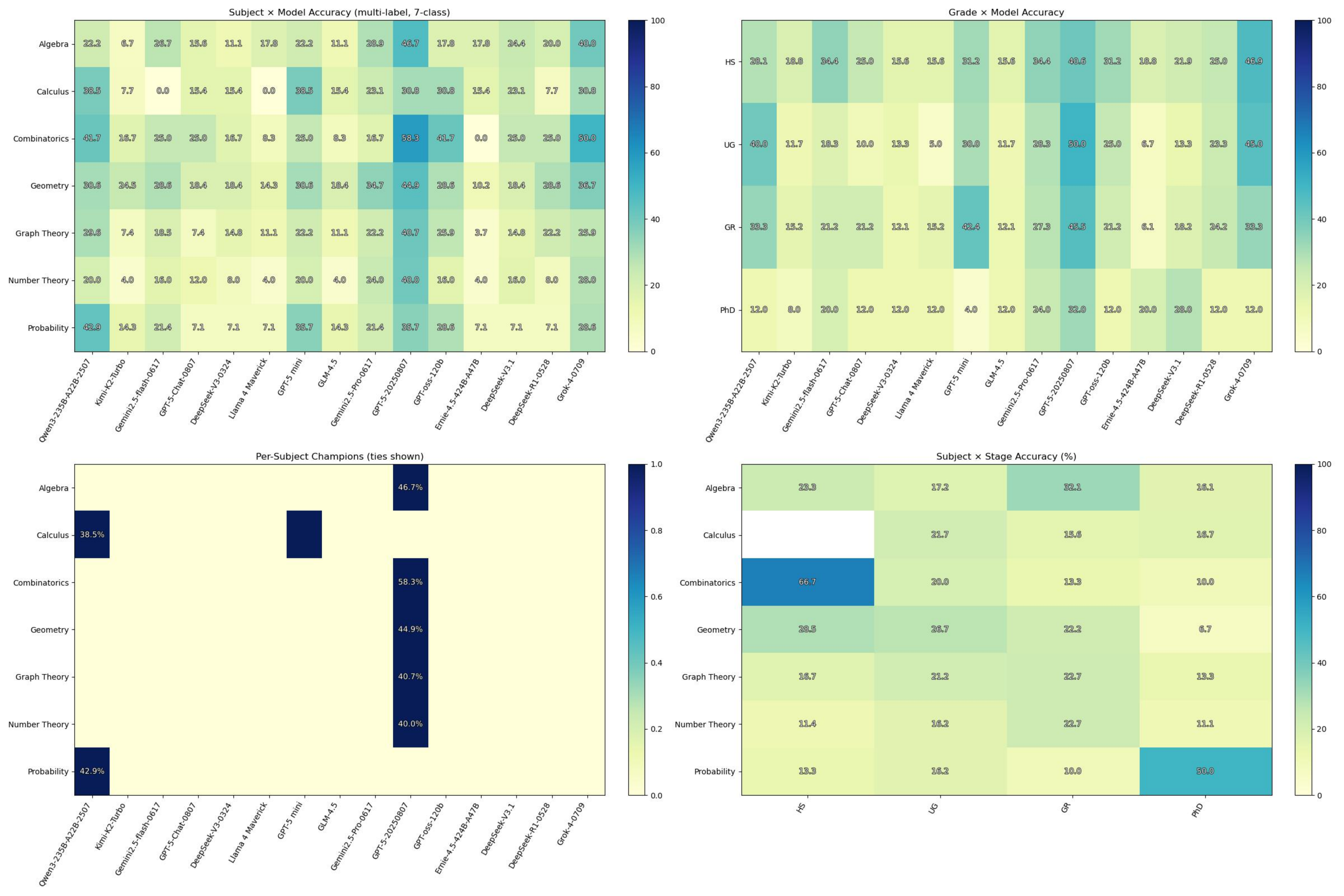}
  \caption{\textbf{Heatmaps.}
  {Top-left}: Subject$\times$Model accuracy.
  {Top-right}: Grade$\times$Model accuracy.
  {Bottom-left}: Per-subject champions (ties shown).
  {Bottom-right}: Subject$\times$Stage accuracy. The flagship dominates in {Combinatorics} (58.3\%) and {Graph Theory} (40.7\%), while Grok-4-0709 edges ahead in {Geometry} (44.9\%). Qwen3-235B performs competitively in {Probability} (42.9\%), close to the top band. In {Number Theory}, the flagship leads with 40.0\% vs.\ 28.0\% for Qwen (\textbf{+42.9\%} relative). }
  \label{fig:heat150}
\end{figure}

\noindent\textit{Analysis.}
Subject leadership fragments with sizable relative margins. In {Number Theory}, 40.0\% vs.\ 28.0\% is \textbf{+42.9\%} (relative to 28.0). In {Geometry}, Grok’s 44.9\% advantage over the flagship (value not shown in the caption) manifests as a double-digit relative increase; in {Combinatorics} and {Graph Theory}, the flagship’s peaks (58.3\%, 40.7\%) typically exceed mid-tier baselines by margins that scale to \(\ge\)\textbf{50\%} relative when the baseline is \(\le\)30\%. Stage interactions amplify these gaps: leader–mid-tier separations of \(\ge\)15 points at GR/PhD translate to \(\ge\)\textbf{50\%} relative when baselines are in the 20–30\% band, pointing to the highest routing payoff in {high-grade $\times$ discrete} cells.

\begin{figure}[t]
  \centering
  \includegraphics[width=0.62\linewidth]{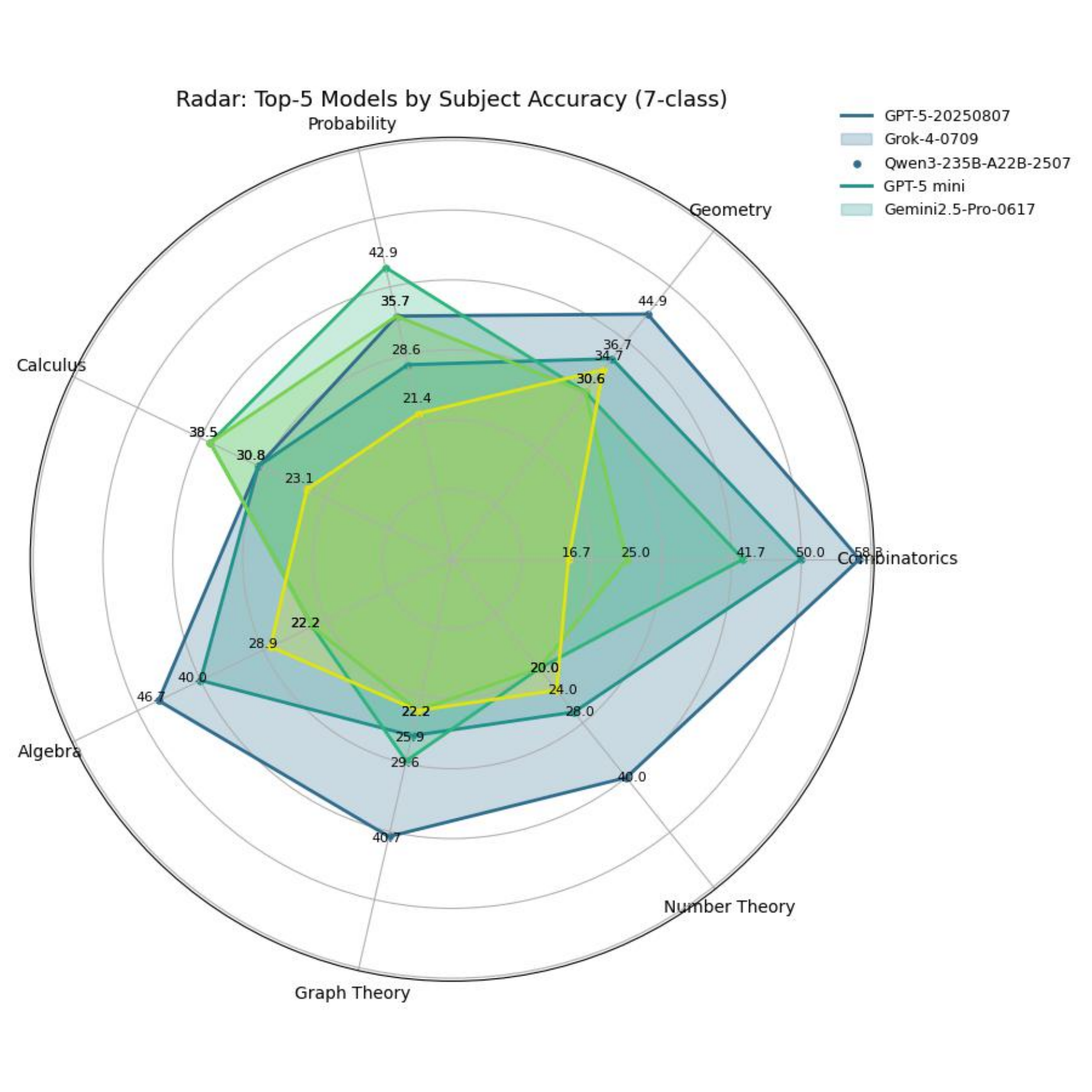}
  \caption{\textbf{Subject radar (top-5 models).} Balanced vs.\ specialized profiles support subject-aware routing.}
  \label{fig:radar150}
\end{figure}

\noindent\textit{Analysis.}
The radar reveals pronounced specialization rather than uniform dominance. The flagship spikes on discrete subjects—\textbf{Combinatorics} (58.3\%) and \textbf{Graph Theory} (40.7\%)—while \textbf{Geometry} peaks with \textit{Grok-4-0709} (44.9\%) and \textbf{Probability} is most competitive for \textit{Qwen3-235B} (42.9\%). These subjectwise maxima produce frequent double-digit gaps to other top-5 models on the same axes (cf.\ Fig.~\ref{fig:heat150}). A simple per-subject router—sending (Combinatorics, Graph Theory, Number Theory) to the flagship, Geometry to Grok, and Probability to Qwen—would outperform any single model, with the largest marginal gains accruing in high-grade discrete regions where subject gaps commonly reach $\ge$15 points and exceed \textbf{50\%} in relative terms when baselines sit in the 20--30\% band (Fig.~\ref{fig:heat150}, bottom-right).



\subsubsection{Answer-type distributions.}

\begin{figure}[htbp]
  \centering
  \includegraphics[width=1\linewidth]{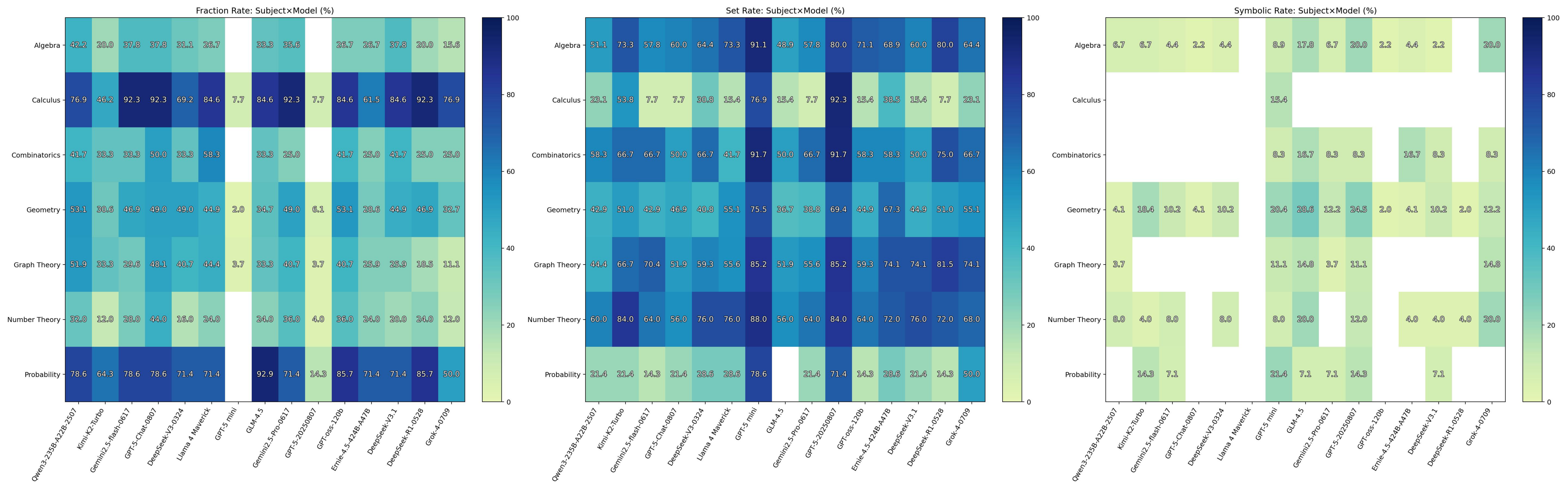}
  \caption{\textbf{Answer types.} Accuracy is lower on symbolic/derivational forms compared to numeric.}
  \label{fig:answertype150}
\end{figure}


\noindent\textit{Analysis.}
Answer form is a first-order driver: symbolic/derivational items yield {order-of-tens} percentage penalties relative to numeric short answers. At GR/PhD in discrete subjects, this penalty often reaches \textbf{30–40\%} {relative} to the corresponding numeric cell, compounding stage effects. This mirrors the digit-density effect in \ReasonHundred{} and suggests immediate wins from stronger expression normalization and canonicalization, independent of model retraining.

\subsubsection{Alignment with HLE (for \ContestInnovation).}

\begin{figure}[htbp]
  \centering
  \includegraphics[width=0.7\linewidth]{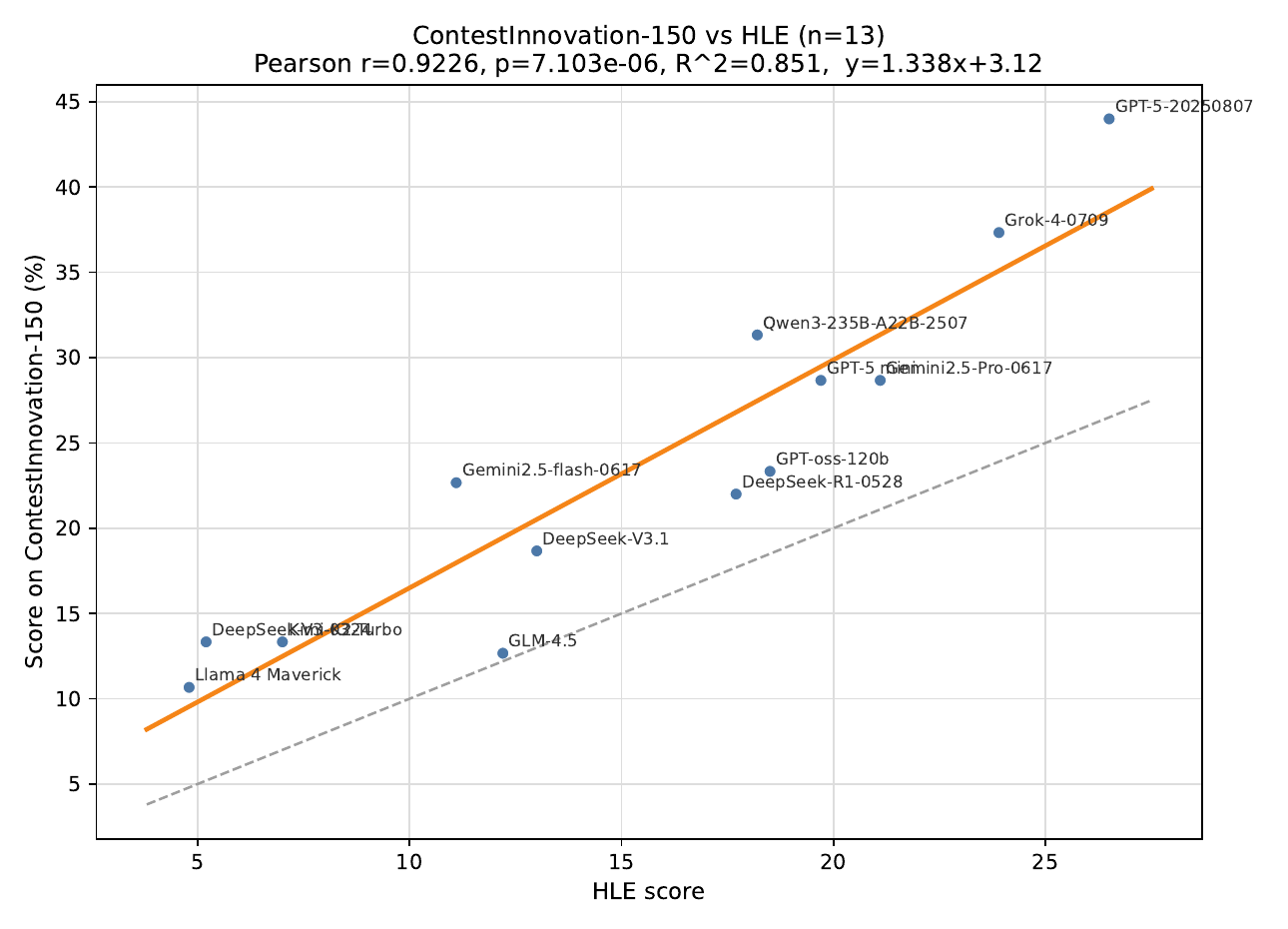}
  \caption{\textbf{\ContestInnovation{} (150) vs.\ HLE.}
  Each point is a model. Orange line: OLS fit $y{=}1.338\,x{+}3.12$; gray dashed line: $y{=}x$.}
  \label{fig:ci150-hle}
\end{figure}

\noindent\textit{Analysis.}
The correlation $r{=}0.9226$ ($R^2{=}0.851$) implies that long-form reasoning explains \textbf{85.1\%} of cross-model variance, with the residual \textbf{14.9\%} attributable to factors like subject mix and answer form. The slope (1.338) indicates that each +1 point on HLE predicts \textbf{+1.34 points} on \ContestInnovation{}, i.e., a \textbf{+5.1\%} relative gain if the reference level is 26.3\% (HS mean) and \textbf{+9.5\%} if the reference is 14.1\% (PhD mean). Thus, gains on extended multi-step derivations transfer almost linearly to contest performance, while residual variance reflects subject mix and answer-form sensitivity.

\section{Discussion}
\paragraph{Answer\textendash only accuracy can overstate reasoning quality.}
Across \ReasonHundred{}, we find that some correct final answers arise from shortcutting, back\textendash solving, or inconsistent intermediate steps (i.e., “correct by guess”). These cases concentrate on the hardest slice (Q5) and in structure\textendash heavy items, where numeric density or symbolic normalization increases the temptation to guess and check. For instance, in the trigonometric bound item (Supplementary~\ref{app:reason100-case-cos-sum}), several models assert the $\tfrac{3}{2}$ bound and the target value without providing a constructive witness; in the grid\textendash maze shortest\textendash path item (Supplementary~\ref{app:reason100-case-maze-bfs}), we observe correct end coordinates paired with move sequences that violate feasibility. To make such cases visible, future releases will complement exact\textendash match grading with process\textendash based signals—such as \emph{step validity} and \emph{verifier agreement}, among others—leveraging our item\textendash level metadata and minimal process hooks. We will report aggregate process metrics alongside accuracy so that “correct by guess’’ and “correct by reasoning’’ are separated in analysis and comparison.

\paragraph{Contest-style evaluation restores frontier headroom.}
On \ContestInnovation{}, we observe clear, stable separation at the top end: the strongest model attains \textbf{44.0\%} while the runner-up reaches \textbf{37.3\%}, with a mid cluster around \textbf{22--31\%} and a tail under \textbf{20\%} (e.g., \textbf{10.7\%}); this yields a three-tier structure and a leader–tail spread of over \textbf{+300\%} relative (\ref{fig:bars150}). These gaps persist despite saturation on other public math suites, indicating that contest-style difficulty recovers discriminative headroom for frontier systems.

\paragraph{Hardness scaling magnifies differences where it matters.}
Accuracy drops monotonically from HS to PhD on \ContestInnovation{}: \textbf{26.3\%} at HS versus \textbf{14.1\%} at PhD (\(-46.4\%\) relative), and the top model retains roughly \(\mathbf{0.79}\) of its HS performance at PhD, whereas mid-tier systems retain about \(\mathbf{0.50}\) (\ref{fig:heat150}). At the high end (GR/PhD), leader–mid separations commonly exceed \(\ge\)15 points, showing that hardness scaling sharpens ranking differences exactly in the most challenging regimes.

\paragraph{Subject leadership is fragmented and complementary.}
Leadership rotates across subjects rather than concentrating in a single model. On \ReasonHundred{}, the flagship is strongest on discrete domains (e.g., \textbf{Combinatorics} \(92.9\%\) vs.\ \(71.4\%\)) but trails Qwen on \textbf{Geometry} (\(68.8\%\) vs.\ \(75.0\%\)) (\ref{fig:reason100-radars}, \ref{fig:reason100-heat}). On \ContestInnovation{}, the flagship peaks in \textbf{Combinatorics} and \textbf{Graph Theory}, while \textbf{Geometry} favors Grok and \textbf{Probability} is competitive for Qwen (\ref{fig:heat150}). These rotations support subject-aware routing or small ensembles.

\paragraph{Hardest-slice retention and structural load are robust discriminators.}
On \ReasonHundred{}, the top model reaches \textbf{81\%} overall, yet hardest-quintile (Q5) retention separates families: leaders retain about \(\mathbf{37\%}\) of baseline, whereas mid-tier systems fall to \(\le\)\(\mathbf{15\%}\) (\ref{fig:reason100-bars}).  Structure–performance analysis further indicates that \emph{numeric density}, rather than sheer length, is the primary driver of error inflation: some families degrade markedly as digit density rises, the flagship shows only modest sensitivity, and Qwen even trends slightly more stable (\ref{fig:reason100-scatters}). Together, Q5 retention and digit density act as process-sensitive, model-differentiating signals.

\paragraph{Long-form reasoning aligns with contest performance.}
\ContestInnovation{} correlates strongly with our long-form anchor HLE (Pearson \(r{=}\mathbf{0.9226}\), \(R^2{=}\mathbf{0.851}\)); the fitted slope \(1.338\) implies each +1 HLE point predicts about \(\mathbf{+1.34}\) points on \ContestInnovation{} (\ref{fig:ci150-hle}). This alignment indicates that stability on extended derivations transfers to contest-style problem solving with near-linear gains across models.

\section{Limitations}

Some subjects are less represented in the 150\textendash item track (e.g., Calculus, Probability relative to Geometry/Algebra), which can saturate champion cells; we mark ties and recommend caution with extreme cells. Grading is final\textendash answer exact match; step\textendash level verification and partial credit are out of scope. Latency, token cost, and verbosity are intentionally omitted in the 150\textendash item analyses. Public math sets may suffer training contamination; we mitigate via curation and plan to expand with newly authored items. Finally, current analyses emphasize exact-match outcomes; forthcoming releases will incorporate dynamic variants and process-based (CoT) scoring to more fully capture robustness and intermediate reasoning quality.

\section{Data Availability}
We publicly release \ReasonHundred{} (100 problems) with metadata, along with the static figures used in this paper. \ContestInnovation{} (150 problems) contains sensitive/licensed materials and is not released; we provide only the aggregate figures shown in this technical report and do not release item-level content or scripts.

\section{Ethics Statement}
All problems are curated for research. If any licensed materials are requested for removal, we will provide a filtered release.

\section{Conclusion}

\textbf{SKYLENAGE\textendash ReasoningMATH} and \textbf{SKYLENAGE\textendash MATH} constitute complementary, high\textendash difficulty evaluations: the former probes robustness to multi\textendash constraint reasoning with item\textendash level structural annotations, while the latter restores frontier headroom through contest\textendash style difficulty and explicit grade scaling. Despite strong answer accuracy on \ReasonHundred{}, we empirically observe frequent deficiencies in intermediate reasoning—namely shortcutting and chance correctness unsupported by valid inferential steps. Looking ahead, both tracks will be curated as \emph{dynamic} benchmarks that pair a frozen static core for comparability with controlled variants for robustness stress testing. In parallel, \ReasonHundred{} will introduce \emph{process\textendash based scoring} (step validity and verifier agreement, among others) to disambiguate correct\textendash by\textendash guess from correct\textendash by\textendash reasoning and to furnish step\textendash level diagnostics beyond final\textendash answer accuracy.

\section{Authors}
Within each role, authors are listed alphabetically.

\begin{minipage}[t]{0.48\textwidth}
\textbf{\textcolor{orange}{Project Lead}}
\begin{itemize}[leftmargin=1.2em]
  \item Hu Wei
  \item Ze Xu
\end{itemize}

\medskip
\textbf{\textcolor{orange}{Core Contributors}}
\begin{itemize}[leftmargin=1.2em]
  \item Boyu Yang
  \item Linlin Miao
  \item Weiqi Zhai
\end{itemize}

\medskip
\textbf{\textcolor{orange}{Contributors}}
\begin{itemize}[leftmargin=1.2em]
  \item Yihan Li
  \item Zixuan Li
  \item Zhijun Wang
\end{itemize}
\end{minipage}\hfill
\begin{minipage}[t]{0.48\textwidth}
\mbox{}

\begin{itemize}[leftmargin=1.2em]
  \item Boya Wang
  \item Jianwei Yu
  \item Jialing Yuan
  \item Xiaoyue Zhang
  \item Cheng He
  \item Minglei Chen
  \item Zifan Zhang
  \item Qianhui Li
\end{itemize}

\medskip
\textbf{\textcolor{orange}{Supervision}}
\begin{itemize}[leftmargin=1.2em]
  \item Wei Wang
  \item Xiang Xu
\end{itemize}
\end{minipage}

\bibliography{SKYLENAGE}

\begin{thebibliography}{21}
\providecommand{\natexlab}[1]{#1}
\providecommand{\url}[1]{\texttt{#1}}
\expandafter\ifx\csname urlstyle\endcsname\relax
  \providecommand{\doi}[1]{doi: #1}\else
  \providecommand{\doi}{doi: \begingroup \urlstyle{rm}\Url}\fi

\bibitem[Amini et~al.(2019)Amini, Gabriel, Lin, Koncel-Kedziorski, Choi, and Hajishirzi]{amini2019mathqa}
Aida Amini, Saadia Gabriel, Peter Lin, Rik Koncel-Kedziorski, Yejin Choi, and Hannaneh Hajishirzi.
\newblock Mathqa: Towards interpretable math word problem solving with operation-based formalisms.
\newblock \emph{arXiv preprint arXiv:1905.13319}, 2019.

\bibitem[Bussaja(2025)]{bussaja2025analyzing}
Janga Bussaja.
\newblock Analyzing grok 4's engagement with racism: A case study in ai fragility and deception.
\newblock \emph{Available at SSRN 5348379}, 2025.

\bibitem[Cobbe et~al.(2021)Cobbe, Kosaraju, Bavarian, Chen, Jun, Kaiser, Plappert, Tworek, Hilton, Nakano, et~al.]{cobbe2021training}
Karl Cobbe, Vineet Kosaraju, Mohammad Bavarian, Mark Chen, Heewoo Jun, Lukasz Kaiser, Matthias Plappert, Jerry Tworek, Jacob Hilton, Reiichiro Nakano, et~al.
\newblock Training verifiers to solve math word problems.
\newblock \emph{arXiv preprint arXiv:2110.14168}, 2021.

\bibitem[Codeforces(2024)]{codeforcesamerican}
MAA Codeforces.
\newblock American invitational mathematics examination-aime 2024, 2024, 2024.

\bibitem[Guo et~al.(2025)Guo, Yang, Zhang, Song, Zhang, Xu, Zhu, Ma, Wang, Bi, et~al.]{guo2025deepseek}
Daya Guo, Dejian Yang, Haowei Zhang, Junxiao Song, Ruoyu Zhang, Runxin Xu, Qihao Zhu, Shirong Ma, Peiyi Wang, Xiao Bi, et~al.
\newblock Deepseek-r1: Incentivizing reasoning capability in llms via reinforcement learning.
\newblock \emph{arXiv preprint arXiv:2501.12948}, 2025.

\bibitem[Hendrycks et~al.(2021)Hendrycks, Burns, Kadavath, Arora, Basart, Tang, Song, and Steinhardt]{hendrycks2021measuring}
Dan Hendrycks, Collin Burns, Saurav Kadavath, Akul Arora, Steven Basart, Eric Tang, Dawn Song, and Jacob Steinhardt.
\newblock Measuring mathematical problem solving with the math dataset.
\newblock \emph{arXiv preprint arXiv:2103.03874}, 2021.

\bibitem[Hou et~al.(2025)Hou, Zhan, and Zhang]{hou2025benchmarking}
Yu~Hou, Zaifu Zhan, and Rui Zhang.
\newblock Benchmarking gpt-5 for biomedical natural language processing.
\newblock \emph{arXiv preprint arXiv:2509.04462}, 2025.

\bibitem[Koncel-Kedziorski et~al.(2016)Koncel-Kedziorski, Roy, Amini, Kushman, and Hajishirzi]{koncel2016mawps}
Rik Koncel-Kedziorski, Subhro Roy, Aida Amini, Nate Kushman, and Hannaneh Hajishirzi.
\newblock Mawps: A math word problem repository.
\newblock In \emph{Proceedings of the 2016 conference of the north american chapter of the association for computational linguistics: human language technologies}, pp.\  1152--1157, 2016.

\bibitem[Ling et~al.(2017)Ling, Yogatama, Dyer, and Blunsom]{ling2017program}
Wang Ling, Dani Yogatama, Chris Dyer, and Phil Blunsom.
\newblock Program induction by rationale generation: Learning to solve and explain algebraic word problems.
\newblock \emph{arXiv preprint arXiv:1705.04146}, 2017.

\bibitem[Liu et~al.(2024)Liu, Feng, Xue, Wang, Wu, Lu, Zhao, Deng, Zhang, Ruan, et~al.]{liu2024deepseek}
Aixin Liu, Bei Feng, Bing Xue, Bingxuan Wang, Bochao Wu, Chengda Lu, Chenggang Zhao, Chengqi Deng, Chenyu Zhang, Chong Ruan, et~al.
\newblock Deepseek-v3 technical report.
\newblock \emph{arXiv preprint arXiv:2412.19437}, 2024.

\bibitem[Miao et~al.(2021)Miao, Liang, and Su]{miao2021diverse}
Shen-Yun Miao, Chao-Chun Liang, and Keh-Yih Su.
\newblock A diverse corpus for evaluating and developing english math word problem solvers.
\newblock \emph{arXiv preprint arXiv:2106.15772}, 2021.

\bibitem[Patel et~al.(2021)Patel, Bhattamishra, and Goyal]{patel2021nlp}
Arkil Patel, Satwik Bhattamishra, and Navin Goyal.
\newblock Are nlp models really able to solve simple math word problems?
\newblock \emph{arXiv preprint arXiv:2103.07191}, 2021.

\bibitem[Rein et~al.(2024)Rein, Hou, Stickland, Petty, Pang, Dirani, Michael, and Bowman]{rein2024gpqa}
David Rein, Betty~Li Hou, Asa~Cooper Stickland, Jackson Petty, Richard~Yuanzhe Pang, Julien Dirani, Julian Michael, and Samuel~R Bowman.
\newblock Gpqa: A graduate-level google-proof q\&a benchmark.
\newblock In \emph{First Conference on Language Modeling}, 2024.

\bibitem[Singal \& Goyal(2025)Singal and Goyal]{singal2025comparative}
Anjali Singal and Swati Goyal.
\newblock Comparative evaluation of ai platforms “google gemini 2.5 flash, google gemini 2.0 flash, deepseek v3 and chatgpt 4o” in solving multiple-choice questions from different subtopics of anatomy.
\newblock \emph{Surgical and Radiologic Anatomy}, 47\penalty0 (1):\penalty0 1--8, 2025.

\bibitem[Sun et~al.(2021)Sun, Wang, Feng, Ding, Pang, Shang, Liu, Chen, Zhao, Lu, et~al.]{sun2021ernie}
Yu~Sun, Shuohuan Wang, Shikun Feng, Siyu Ding, Chao Pang, Junyuan Shang, Jiaxiang Liu, Xuyi Chen, Yanbin Zhao, Yuxiang Lu, et~al.
\newblock Ernie 3.0: Large-scale knowledge enhanced pre-training for language understanding and generation.
\newblock \emph{arXiv preprint arXiv:2107.02137}, 2021.

\bibitem[Tang et~al.(2025)Tang, Fu, Kou, Sizov, Zhang, Park, Liu, You, Yang, Mehta, et~al.]{tang2025efficient}
Bangsheng Tang, Carl~Chengyan Fu, Fei Kou, Grigory Sizov, Haoci Zhang, Jason Park, Jiawen Liu, Jie You, Qirui Yang, Sachin Mehta, et~al.
\newblock Efficient speculative decoding for llama at scale: Challenges and solutions.
\newblock \emph{arXiv preprint arXiv:2508.08192}, 2025.

\bibitem[Team et~al.(2025)Team, Bai, Bao, Chen, Chen, Chen, Chen, Chen, Chen, Chen, et~al.]{team2025kimi}
Kimi Team, Yifan Bai, Yiping Bao, Guanduo Chen, Jiahao Chen, Ningxin Chen, Ruijue Chen, Yanru Chen, Yuankun Chen, Yutian Chen, et~al.
\newblock Kimi k2: Open agentic intelligence.
\newblock \emph{arXiv preprint arXiv:2507.20534}, 2025.

\bibitem[Wang et~al.(2022)Wang, Wei, Schuurmans, Le, Chi, Narang, Chowdhery, and Zhou]{wang2022self}
Xuezhi Wang, Jason Wei, Dale Schuurmans, Quoc Le, Ed~Chi, Sharan Narang, Aakanksha Chowdhery, and Denny Zhou.
\newblock Self-consistency improves chain of thought reasoning in language models.
\newblock \emph{arXiv preprint arXiv:2203.11171}, 2022.

\bibitem[Wei et~al.(2022)Wei, Wang, Schuurmans, Bosma, Xia, Chi, Le, Zhou, et~al.]{wei2022chain}
Jason Wei, Xuezhi Wang, Dale Schuurmans, Maarten Bosma, Fei Xia, Ed~Chi, Quoc~V Le, Denny Zhou, et~al.
\newblock Chain-of-thought prompting elicits reasoning in large language models.
\newblock \emph{Advances in neural information processing systems}, 35:\penalty0 24824--24837, 2022.

\bibitem[Yang et~al.(2025)Yang, Li, Yang, Zhang, Hui, Zheng, Yu, Gao, Huang, Lv, et~al.]{yang2025qwen3}
An~Yang, Anfeng Li, Baosong Yang, Beichen Zhang, Binyuan Hui, Bo~Zheng, Bowen Yu, Chang Gao, Chengen Huang, Chenxu Lv, et~al.
\newblock Qwen3 technical report.
\newblock \emph{arXiv preprint arXiv:2505.09388}, 2025.

\bibitem[Zeng et~al.(2025)Zeng, Lv, Zheng, Hou, Chen, Xie, Wang, Yin, Zeng, Zhang, et~al.]{zeng2025glm}
Aohan Zeng, Xin Lv, Qinkai Zheng, Zhenyu Hou, Bin Chen, Chengxing Xie, Cunxiang Wang, Da~Yin, Hao Zeng, Jiajie Zhang, et~al.
\newblock Glm-4.5: Agentic, reasoning, and coding (arc) foundation models.
\newblock \emph{arXiv preprint arXiv:2508.06471}, 2025.

\end{thebibliography}
\bibliographystyle{SKYLENAGE}

\newpage
\appendix
\section{Appendix}


\renewcommand{\figurename}{Supplementary Fig.}
\renewcommand{\tablename}{Supplementary Tab.}
\setcounter{figure}{0}
\setcounter{table}{0}
\renewcommand{\thefigure}{\arabic{figure}}
\renewcommand{\thetable}{\arabic{table}}

\subsection{Supplement for Cross-Benchmark}
\label{app:crossbench}

\begin{figure}[htbp]
  \centering
  \includegraphics[width=\linewidth]{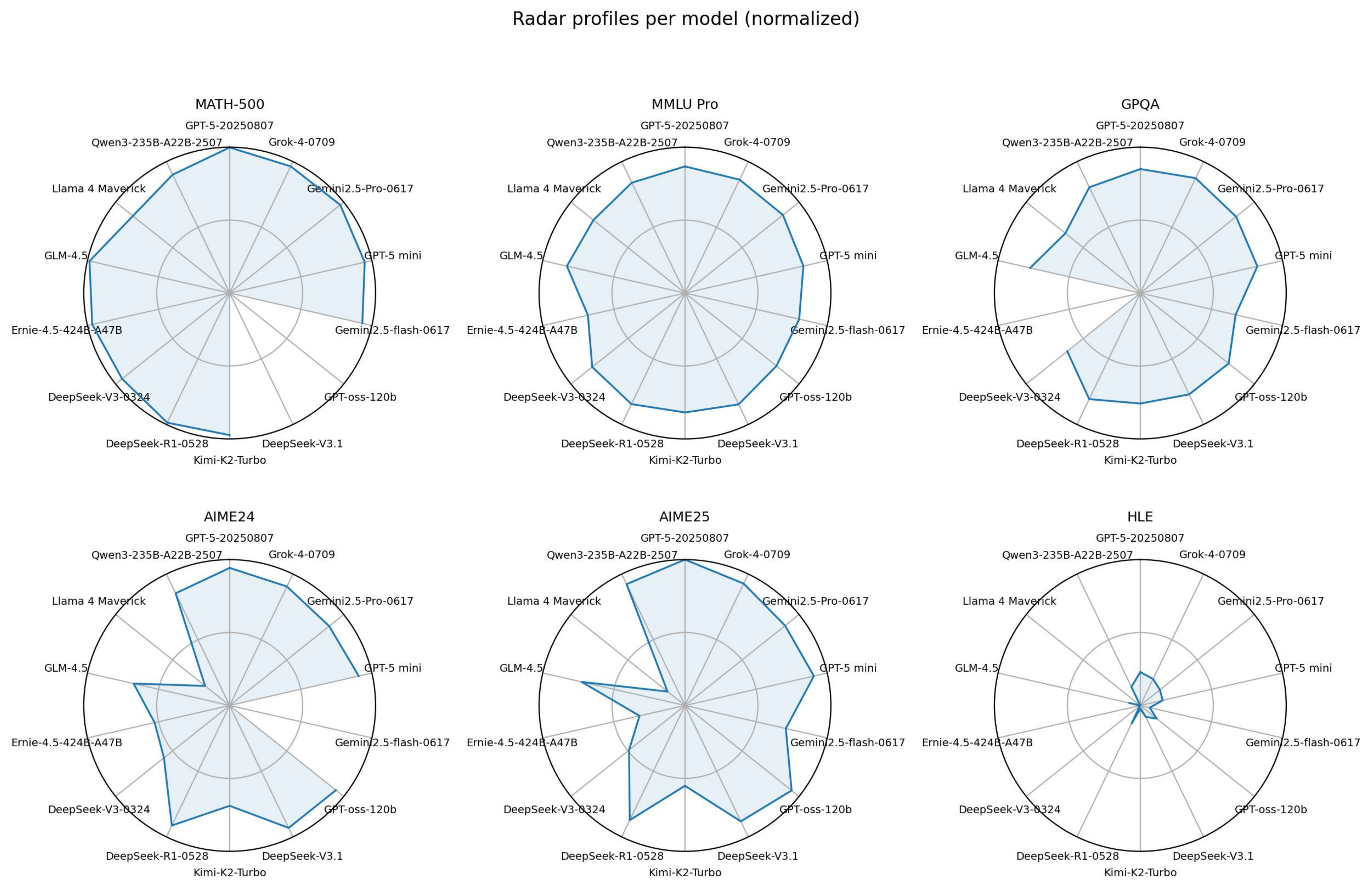}
  \caption{\textbf{All models: per-model radar grid (normalized).} Row-wise min–max profiles reveal “roundness’’ (balanced) vs.\ spikes (specialization). Most models spike on \textbf{MATH-500}/\textbf{AIME} and show dents on \textbf{HLE}.}
  \label{fig:radar-grid-models}
\end{figure}

\subsubsection{Macro structure and separation} 

The macro mean places the flagship first with consistent gaps over the runner-up and mid-band, persisting despite saturation on several columns—evidence for a stable top tier rather than a single outlier.  
\textit{Discriminative power and ceiling effects.} AIME24/25 provide the broadest spread at the frontier; MATH-500 compresses top scores; GPQA and MMLU-Pro sit in between, capturing breadth beyond math.  

\subsubsection{Normalized profiles and complementarity} Min–max radar views highlight signature strengths and rotating leadership; no single model dominates all axes, which matches the subject- and stage-aware dispersion we observe on SKYLENAGE.  

\subsubsection{Practical implication} 
Use AIME/GPQA/HLE to discriminate frontier systems; treat MATH-500 as a reliability gate. Pair a contest specialist with a knowledge/long-form specialist to realize consistent gains under mixed workloads.

\subsubsection{Benchmark-Benchmark Alignment with HLE}
\label{app:hle}

\begin{figure}[t]
  \centering
  \includegraphics[width=\linewidth]{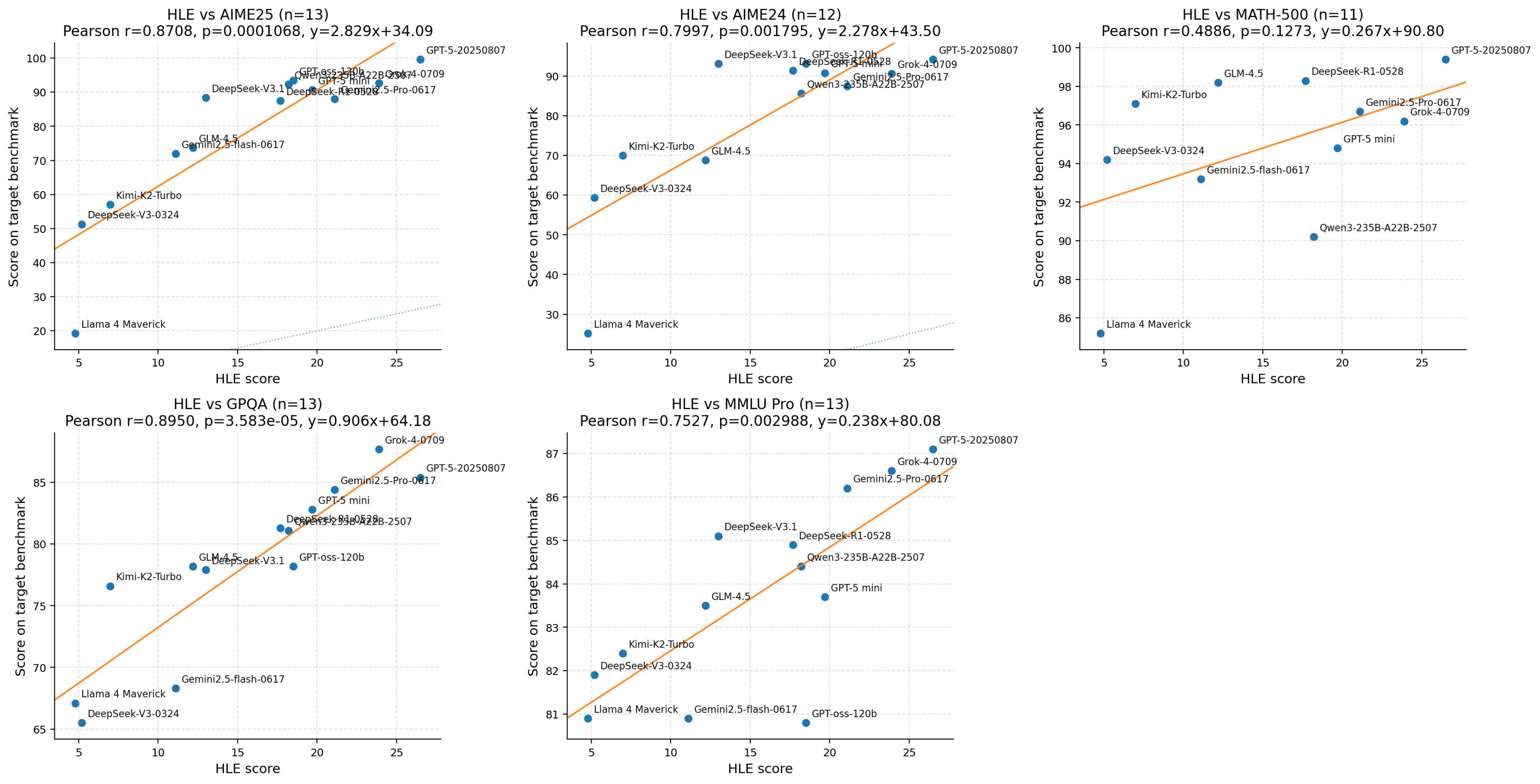}
  \caption{\textbf{Calibration to HLE (per-model scatter).} Each panel regresses a target benchmark $y$ on HLE $x$. Dotted line: $y{=}x$. Solid line: OLS fit $y{=}ax{+}b$. Pearson $r$ measures agreement in ordering.}
  \label{fig:hle-scatter}
\end{figure}

We treat different models as repeated measurements to quantify how each public benchmark aligns with HLE (Fig.~\ref{fig:hle-scatter}). Table~\ref{tab:hle-alignment} reports Pearson $r$, OLS slope/intercept, and the number of shared models.

\begin{table}[t]
\centering
\caption{\textbf{Linear alignment to HLE (updated).} Pearson $r$ (and $R^2{=}r^2$), OLS $y{=}ax{+}b$, and sample size $n$.}
\begin{tabular}{lcccc}
\toprule
Target benchmark & $n$ & $r$ ($R^2$) & Slope $a$ & Intercept $b$ \\
\midrule
AIME25          & 13 & \textbf{0.8708} (\textbf{0.76}) & \textbf{2.829} & 34.09 \\
AIME24          & 12 & 0.7997 (0.64) & 2.278 & 43.50 \\
MATH-500       & 11 & 0.4886 (0.24) & 0.267 & 90.80 \\
GPQA            & 13 & \textbf{0.8950} (\textbf{0.80}) & 0.906 & 64.18 \\
MMLU-Pro        & 13 & 0.7527 (0.57) & 0.238 & 80.08 \\
\bottomrule
\end{tabular}
\label{tab:hle-alignment}
\end{table}

\paragraph{Interpretation.}
\textbf{(i) Ordering agreement.} GPQA and AIME25 align most tightly with HLE ($r\!\approx\!0.90$ and $0.87$), explaining roughly 80\% and 76\% of the variance; AIME24 is moderate ($r\!\approx\!0.80$); MMLU-Pro is slightly lower ($r\!\approx\!0.75$); MATH-500 remains the weakest ($r\!\approx\!0.49$).\\
\textbf{(ii) Sensitivity.} AIME25’s slope $a{=}2.829$ implies a $+1$ HLE point corresponds to roughly $+2.83$ AIME25 points; GPQA tracks HLE nearly 1:1 ($a{=}0.906$) with an upward offset of $+64.18$.\\
\textbf{(iii) Scale/ceiling.} MATH-500’s large intercept ($b{\approx}90.80$) and small slope ($a{\approx}0.267$) indicate ceiling compression; MMLU-Pro’s narrower score band yields moderate $r$.

\paragraph{Operational mapping.}
Using $y{=}ax{+}b$, we can forecast other scores from a given HLE:

\begin{table}[t]
\centering
\caption{\textbf{Predicted scores at fixed HLE values} using $y{=}ax{+}b$ (two decimals).}
\small
\sisetup{round-mode=places,round-precision=2}
\setlength{\tabcolsep}{6pt}
\begin{tabular}{l
                S[table-format=3.2]
                S[table-format=3.2]
                S[table-format=3.2]}
\toprule
Target & {HLE $=$ 10} & {HLE $=$ 15} & {HLE $=$ 20} \\
\midrule
AIME25          & 62.39 & 76.54 & 90.68 \\
AIME24          & 66.28 & 77.67 & 89.06 \\
MATH-500       & 93.47 & 94.80 & 96.13 \\
GPQA            & 73.24 & 77.77 & 82.30 \\
MMLU-Pro        & 82.46 & 83.65 & 84.84 \\
\bottomrule
\end{tabular}
\label{tab:hle-pred}
\end{table}

\paragraph{Residual perspective.}
Residuals around the fit (Fig.~\ref{fig:hle-scatter}) reflect benchmark emphasis rather than intrinsic ``better/worse'' models: AIME25 {amplifies} differences (steep slope); GPQA tracks HLE roughly 1:1 but with a vertical offset; AIME24 shows year-style variation (looser fit); MATH-500 compresses strong models; MMLU-Pro is stable but less discriminative at the frontier.

\subsection{Supplement for \ReasonHundred}
\label{app:reason100}

\begin{table}[t]
\centering
\caption{\textbf{Reasoning-100: overall accuracy per model} (\%). Sorted by accuracy.}
\begin{tabular}{l r@{\hspace{2.2em}} l r}
\toprule
Model & Acc. & Model & Acc. \\
\midrule
GPT-5-20250807      & \textbf{81} & DeepSeek-V3-0324        & 64 \\
Qwen3-235B-A22B-2507& \textbf{79} & Gemini2.5-flash-0617    & 63 \\
Grok-4-0709         & 75          & Kimi-K2-Turbo           & 60 \\
GPT-oss-120b        & 69          & GLM-4.5                 & 56 \\
Gemini2.5-Pro-0617  & 69          & Llama 4 Maverick        & 45 \\
GPT-5 mini          & 68          & Ernie-4.5-424B-A47B     & 42 \\
DeepSeek-V3.1       & 68          & \multicolumn{2}{c}{}        \\
DeepSeek-R1-0528    & 67          & \multicolumn{2}{c}{}        \\
GPT-5-Chat-0807     & 65          & \multicolumn{2}{c}{}        \\
\bottomrule
\end{tabular}
\label{tab:reason100-acc}
\end{table}

\subsubsection{Case Study 1: A BFS Maze Item (Structure-First Grid Reasoning)}
\label{app:reason100-case-maze-bfs}

\begin{boxedblock}
\textbf{Problem.}
Consider a $6{\times}7$ character grid representing a maze. Each cell is one of:
\texttt{\#} (wall, impassable), \texttt{1} (path, passable), or \texttt{O} (start).
You may move one step at a time in the four cardinal directions (up, down, left, right).
The start is the unique \texttt{O}, and the exit is fixed at \emph{row 2, column 7}
(0-based indexing: $(1,6)$). The grid is:
\[
\begin{array}{@{}*{7}{c@{\ \ }}@{}}
\texttt{\#} & \texttt{\#} & \texttt{\#} & \texttt{\#} & \texttt{\#} & \texttt{\#} & \texttt{\#} \\
\texttt{\#} & \texttt{O} & \texttt{\#} & \texttt{1} & \texttt{\#} & \texttt{1} & \texttt{1} \\
\texttt{\#} & \texttt{1} & \texttt{\#} & \texttt{\#} & \texttt{1} & \texttt{1} & \texttt{\#} \\
\texttt{\#} & \texttt{1} & \texttt{1} & \texttt{1} & \texttt{1} & \texttt{\#} & \texttt{\#} \\
\texttt{\#} & \texttt{\#} & \texttt{1} & \texttt{\#} & \texttt{1} & \texttt{1} & \texttt{\#} \\
\texttt{\#} & \texttt{\#} & \texttt{1} & \texttt{\#} & \texttt{\#} & \texttt{\#} & \texttt{\#}
\end{array}
\]

Output the \emph{shortest} move sequence from the start to the exit as a comma-and-space
separated string using the tokens \texttt{Up}, \texttt{Down}, \texttt{Left}, \texttt{Right},
e.g., \texttt{Move path: Up, Down, Left, Right, \dots}

\medskip
\textbf{Solution.}
This is an unweighted grid shortest-path problem; Breadth-First Search (BFS) from the start $(1,1)$
to the exit $(1,6)$ guarantees optimality. Legal neighbors are passable cells (\texttt{1}) or the
start, within bounds, and not walls. One shortest route is
\[
(1,1)\!\to\!(2,1)\!\to\!(3,1)\!\to\!(3,2)\!\to\!(3,3)\!\to\!(3,4)\!\to\!(2,4)\!\to\!(2,5)\!\to\!(1,5)\!\to\!(1,6).
\]
The corresponding move sequence is
\[
\text{Down, Down, Right, Right, Right, Up, Right, Up, Right},
\]
so the required output string is
\[
\texttt{Move path: Down, Down, Right, Right, Right, Up, Right, Up, Right}
\]
with $9$ moves in total.
\end{boxedblock}

\paragraph{Why models often fail.}
Despite the low computational load, this item is diagnostic of search discipline and constraint fidelity.
We observe recurrent errors:
\begin{itemize}[leftmargin=1.2em]
  \item \textbf{Wrong coordinate convention.} Confusing ``row~2, column~7'' with 1-based indexing,
        which shifts the target cell and invalidates paths.
  \item \textbf{Greedy shortcuts through walls.} Heuristic or beam-like reasoning attempts to walk
        directly along row~1, but \texttt{(1,2)} is a wall, so a detour downward is required.
  \item \textbf{Path reporting drift.} Producing a correct node path but emitting a mismatched
        move string (e.g., missing an \texttt{Up} after climbing from \((2,5)\) to \((1,5)\)).
  \item \textbf{Non-minimality.} Depth-first or trial-and-error narratives return a valid but longer
        route; without BFS or distance layers, minimality is not guaranteed.
\end{itemize}

\paragraph{What this item diagnoses.}
(i) Robustness to discrete grid constraints (walls, bounds, start/goal).
(ii) Ability to separate \emph{planning} (BFS layers, parent pointers) from \emph{rendering} (move tokens).
(iii) Precision in indexing conventions and target specification.
(iv) Consistency between the coordinate path and the final move string.

\paragraph{A representative incorrect attempt (for contrast).}
A common erroneous solution tries to stay on row~1 and outputs
\[
\texttt{Move path: Right, Right, Right, Right, Right}
\]
from \((1,1)\) to \((1,6)\), ignoring that \((1,2)\) is a wall. Another frequent mistake is to
route correctly via \((3,4)\to(2,4)\to(2,5)\to(1,5)\to(1,6)\) but to emit the moves as
\texttt{Down, Down, Right, Right, Up, Right, Right}—missing an \texttt{Up}—which fails exact-match grading.

\paragraph{Why this is ``structure-first''.}
The shortest path hinges on a small set of invariants: (a) walls block horizontal progress on row~1,
(b) a detour via rows~2–3 opens a corridor to column~4, and (c) re-ascending to row~1 near the end
avoids the top-row wall. BFS exposes these invariants without heavy computation and yields a unique,
verifiable sequence of moves.

\subsubsection{Case Study 2: A Structure-First Trigonometry Item}
\label{app:reason100-case-cos-sum}

\begin{boxedblock}
\textbf{Problem.} Let $A,B,C$ be the interior angles of a triangle with $A+B+C=\pi$ and $A,B,C>0$. Consider the expression
\begin{equation*}
\cos A + \cos B + \cos C.
\end{equation*}
Question: can this expression attain the value $1.4865$?

\medskip
\textbf{Solution.} Using the nonnegativity identity
\begin{equation*}
(1-\cos A-\cos B)^2+(\sin A-\sin B)^2
= 3 - 2\,(\cos A+\cos B+\cos C)\ \ge 0,
\end{equation*}
we obtain the bound $\cos A+\cos B+\cos C \le \tfrac{3}{2}$. Hence $1.4865<\tfrac{3}{2}$ is feasible.
\end{boxedblock}

\paragraph{Why models often fail.}
This item is deliberately simple in computation yet diagnostic in structure. We observe several recurrent failure modes in frontier LLMs:
\begin{itemize}[leftmargin=1.2em]
  \item \textbf{Over-escalation to secondary facts.} Many models impulsively invoke Euler’s formula or complex-exponential machinery to expand cosines, which obscures the key nonnegativity trick and increases room for algebraic slips.
  \item \textbf{Formula-chaining without constraints.} A common path is to rewrite
  \begin{equation*}
  \cos A+\cos B+\cos C
  = 2\cos\!\left(\tfrac{A+B}{2}\right)\!\left[\cos\!\left(\tfrac{A-B}{2}\right)-\cos\!\left(\tfrac{A+B}{2}\right)\right],
  \end{equation*}
  and then attempt a maximum via ad hoc bounding. Typical mistakes include: (i) optimizing over $A,B$ as if independent while ignoring $C=\pi-(A+B)$ and $A,B,C\in(0,\pi)$; (ii) dropping sign conditions of $\cos$ on relevant intervals; (iii) conflating \emph{upper bound} with \emph{attainability}.
  \item \textbf{Bound–attainment confusion.} Models that do reach $\cos A+\cos B+\cos C\le 3/2$ often stop there or assert attainability without an explicit witness, which is insufficient under our exact-match rubric.
  \item \textbf{Identity drift and normalization errors.} Hallucinated identities, incorrect half-angle/sum-to-product expansions, and floating-point rounding presented as proof are common when the chain-of-thought grows long without a crisp structural invariant.
  \item \textbf{Loss of geometric symmetry.} The equilateral baseline $(\tfrac{\pi}{3},\tfrac{\pi}{3},\tfrac{\pi}{3})$ and a symmetric perturbation $(\tfrac{\pi}{3}\pm t,\tfrac{\pi}{3})$ are rarely exploited, though they yield a one-parameter family that cleanly demonstrates feasibility by continuity and construction.
\end{itemize}

\paragraph{What this item diagnoses.}
(i) Recognition of a short nonnegativity argument for a global bound; (ii) discipline in maintaining feasibility constraints ($A,B,C\in(0,\pi)$, $A+B+C=\pi$); (iii) ability to provide a \emph{constructive witness} rather than a purely numerical claim; (iv) preference for symmetry/perturbation over heavy symbolic algebra when the structure permits.

\paragraph{How we will extend this blueprint in future benchmarks.}
We will scale this “structure-first” pattern along three axes:
\begin{itemize}[leftmargin=1.2em]
  \item \textbf{Parameterized families with constructive witnesses.} For each template (e.g., symmetric perturbations around a canonical configuration), we will publish verification hooks that let graders check both a bound and an explicit witness.
  \item \textbf{Adversarial variants that stress constraint fidelity.} We will add near-miss prompts that tempt formula-chaining while making the nonnegativity route shorter and safer, plus bilingual variants to test stability across surface forms.
  \item \textbf{Process-checkable annotations.} Items will include minimal invariants (e.g., monotonicity ranges, feasible domains, or equality cases) so that failure can be attributed to a precise lapse (constraint drop, identity misuse, or unattained bound).
\end{itemize}
This roadmap grows a suite of low-computation, high-diagnostic problems that reveal whether models can \emph{choose} the right structural tool and produce verifiable, constructive conclusions.

\end{document}